\documentclass{article}

\usepackage{microtype}
\usepackage{graphicx}
\usepackage{subfigure}
\usepackage{booktabs} % for professional tables

\usepackage{hyperref}

\usepackage[preprint]{icml2026}

\usepackage{amsmath}
\usepackage{amssymb}
\usepackage{mathtools}
\usepackage{amsthm}
\usepackage{multirow}
\usepackage{bbding}
\usepackage{makecell}

\usepackage{booktabs}
\usepackage{graphicx}
\usepackage{bbm}
\usepackage[capitalize,noabbrev]{cleveref}

\theoremstyle{plain}

\theoremstyle{definition}

\theoremstyle{remark}

\usepackage[disable,textsize=tiny]{todonotes}
\usepackage[textsize=tiny]{todonotes}

\icmltitlerunning{Semantic One-Dimensional Tokenizer for Image Reconstruction and Generation}

\begin{document}

\twocolumn[
\icmltitle{Semantic One-Dimensional Tokenizer for Image Reconstruction and Generation}

% It is OKAY to include author information, even for blind
% submissions: the style file will automatically remove it for you
% unless you've provided the [accepted] option to the icml2026
% package.

% List of affiliations: The first argument should be a (short)
% identifier you will use later to specify author affiliations
% Academic affiliations should list Department, University, City, Region, Country
% Industry affiliations should list Company, City, Region, Country

% You can specify symbols, otherwise they are numbered in order.
% Ideally, you should not use this facility. Affiliations will be numbered
% in order of appearance and this is the preferred way.
\icmlsetsymbol{equal}{*}

\begin{icmlauthorlist}
\icmlauthor{Yunpeng Qu}{equal,sch}
\icmlauthor{Kaidong Zhang}{equal,comp}
\icmlauthor{Yukang Ding}{comp}
\icmlauthor{Ying Chen}{comp}
\icmlauthor{Jian Wang}{sch}
\end{icmlauthorlist}

\icmlaffiliation{sch}{Tsinghua University, Beijing, China}
\icmlaffiliation{comp}{Alibaba Group, Beijing, China}
%\icmlaffiliation{sch}{School of ZZZ, Institute of WWW, Location, Country}

\icmlcorrespondingauthor{Yukang Ding}{dingyukang.dyk@taobao.com}
\icmlcorrespondingauthor{Jian Wang}{jian-wang@tsinghua.edu.cn}

% You may provide any keywords that you
% find helpful for describing your paper; these are used to populate
% the "keywords" metadata in the PDF but will not be shown in the document
\icmlkeywords{Machine Learning, ICML}

\vskip 0.3in
]

% this must go after the closing bracket ] following \twocolumn[ ...

% This command actually creates the footnote in the first column
% listing the affiliations and the copyright notice.
% The command takes one argument, which is text to display at the start of the footnote.
% The \icmlEqualContribution command is standard text for equal contribution.
% Remove it (just {}) if you do not need this facility.

%\printAffiliationsAndNotice{}  % leave blank if no need to mention equal contribution
\printAffiliationsAndNotice{\icmlEqualContribution} % otherwise use the standard text.

\begin{abstract}
Visual generative models based on latent space have achieved great success, underscoring the significance of visual tokenization.
Mapping images to latents boosts efficiency and enables multimodal alignment for scaling up in downstream tasks.
Existing visual tokenizers primarily map images into fixed 2D spatial grids and focus on pixel-level restoration, which hinders the capture of representations with compact global semantics.
To address these issues, we propose \textbf{SemTok}, a semantic one-dimensional tokenizer that compresses 2D images into 1D discrete tokens with high-level semantics.
SemTok sets a new state-of-the-art in image reconstruction, achieving superior fidelity with a remarkably compact token representation. This is achieved via a synergistic framework with three key innovations: a 2D-to-1D tokenization scheme, a semantic alignment constraint, and a two-stage generative training strategy.
%Leveraging a 2D-to-1D conversion framework, semantic alignment constraint, and two-stage generative training, SemTok improves tokenization form, constraints, and training strategy, achieving SOTA image reconstruction performance with a compact compression ratio.
Building on SemTok, we construct a masked autoregressive generation framework, which yields notable improvements in downstream image generation tasks. 
Experiments confirm the effectiveness of our semantic 1D tokenization.
Our code will be open-sourced.
% Existing visual tokenizers primarily map images into fixed 2D spatial grid structures, which hinders the capture of global semantics. 
% Additionally, these tokenizers focus on pixel-level texture restoration rather than the representation of global high-level semantics.
%delivers state-of-the-art performance with a compact compression ratio for image reconstruction.

\end{abstract}

\section{Introduction}
\label{sec:intro}
Visual generative models based on diffusion models \cite{rombach2022high, ho2020denoising, chen2023pixart, dhariwal2021diffusion} or autoregressive modeling (AR) \cite{chang2022maskgit, tian2024visual, sun2024autoregressive, xie2024show} have achieved significant success and are widely used in creative applications.
%Visual generative models based on diffusion models \cite{rombach2022high, ho2020denoising, chen2023pixart, dhariwal2021diffusion} or autoregressive modeling (AR) \cite{chang2022maskgit, tian2024visual, sun2024autoregressive, xie2024show} are widely used in creative applications.
These models typically operate in a latent space rather than pixel space, using a tokenizer to map images to continuous \cite{kingma2013auto} or discretized \cite{esser2021taming} representations.
% These models typically operate in a latent space rather than pixel space, using a tokenizer to map raw pixels to continuous \cite{kingma2013auto} or discretized \cite{esser2021taming} representations.
%Existing generative models are often based on the latent space of images rather than pixel space, which means they require an encoder and decoder to transform raw pixels into latent representations, including continuous \cite{kingma2013auto} or discretized vectors \cite{esser2021taming}.
Latent representations are more compact and expressive than raw pixels, simplifying complexity and facilitating integration with multimodal data \cite{yu2024image}.
%The latent space provides a more compact and expressive image representation than pixel space, accelerating training and inference processes, and facilitating integration with multimodal data (\textit{e.g.}, text, audio) \cite{yu2024image}.

While image tokenizers like VQGAN \cite{esser2021taming} achieve great success in visual generation and understanding tasks, they adhere to the standard spatial structure assumption, representing images as two-dimensional (2D) spatial grids \cite{wang2025selftok}.
The rigid spatial positioning of grid-based tokens biases them towards capturing local semantics, which hinders the formation of a compact, globally-aware image representation due to inherent local redundancies.
%Tokens corresponding to grid points are constrained by their spatial positions, 
%which limits them to convey local semantics rather than global semantics and thus impedes more compact image representations, owing to the inherent redundancy that tends to exist among local semantics.
Such 2D representations also limit their applicability to downstream tasks, as tokens are typically flattened into one-dimensional (1D) form in generation frameworks, making it more complex to model spatial structures.
Moreover, the bidirectional spatial correlations of 2D tokens cannot model causal dependencies as in language \cite{tian2024visual}.
%limiting them to convey local semantics rather than global semantics, thus hindering more compressed image representation.
%hindering the perception of global semantics to obtain more compressed image representations.
%The 2D token map is typically flattened to a one-dimensional (1D) form in transformer-based generation frameworks, complicating spatial structure modeling.
%Additionally, the bidirectional correlations of the spatial tokens cannot model causal dependencies as effectively as in language.

Besides, existing visual tokenizers are trained on image reconstruction tasks, with losses aimed at optimizing pixel-level reconstruction errors \cite{van2017neural}.
This causes the model to prioritize low-level texture reconstruction over high-level semantic representation, which limits both information density and the low-entropy optimization of the distribution.
%This results in the model focusing more on low-level texture reconstruction rather than global semantic restoration.
While the integration of LPIPS \cite{zhang2018unreasonable} and GAN \cite{goodfellow2020generative} losses enhances fidelity, such losses backpropagate indirectly from the decoder to the encoder, lacking the direct constraint on the encoder's inherent representation capability, which is crucial for integration with LLM-based understanding and generative frameworks \cite{wang2024emu3, ge2024seed}.

%Above limitations pose a critical question: \textit{what tokenizer enables more compact and precise semantic representations?}
The aforementioned limitations underscore the need for a more advanced tokenizer, raising a key research question: \textit{how can we design a tokenizer that yields more compact and semantically richer representations?}
We believe that compact representations stem from the low-entropy optimization of the latent space, as compressed tokens should exhibit more clustered distributions to capture more discriminative and representative semantics.
This motivates us to develop a more compact tokenization scheme from three aspects:
(1) \textbf{\textit{Tokenization form}}: Inspired by prior works \cite{yu2024image, bachmann2025flextok, wang2025selftok} that compress images into 1D sequences, we highlight the value of 1D visual representations, which enable tokens to prioritize critical global semantics while eliminating redundancy among local information.
(2) \textbf{\textit{Tokenization constraint}}: It is necessary to impose explicit distributional constraints on the encoder side, so as to drive compressed tokens toward more distinct and concentrated semantic representations.
(3) \textbf{\textit{Tokenization training paradigm}}: We aim at exploring a more diverse latent space via a generative training paradigm to avoid potential distribution collapse of the latent space from pixel-level reconstruction.

%Many related works (\textit{e.g.}, TiTok \cite{yu2024image}, FlexTok \cite{bachmann2025flextok} and SelfTok \cite{wang2025selftok}) represent images as 1D sequences rather than 2D maps, enabling improved compression efficiency while maintaining competitive generation quality.
%Inspired by this progress, we highlight the value of 1D visual representations, which enable tokens to prioritize critical global semantics for more concise image encoding. 
% However, these works still rely on standard end-to-end image reconstruction optimization, implying a lack of explicit constraint on the accuracy of semantic representations. 
% Furthermore, they often involve complex training pipelines and rely on pretrained 2D tokenizers for distillation or prior provision, which may obscure the semantic preservation advantages inherently offered by 1D encoding.

In this paper, we introduce a \textbf{Semantic one-dimensional Tokenizer (SemTok)}, which compresses the image into a 1D discrete sequence imbued with high-level semantics, and reconstructs the image via a generative decoder.
To realize the 1D tokenization form, we leverage the \textit{2D-to-1D conversion framework} based on MMDiT \cite{esser2024scaling} to encode the semantics of 2D images into compact 1D tokens.
To concentrate the latent distribution, we introduce a \textit{semantic alignment constraint} to directly supervise the encoder in generating clustered tokens aligned with multimodal inputs, boosting downstream representational capacity.
%thereby enhancing its representational capacity in downstream tasks.
To explore a more diverse latent space, we propose a \textit{two-stage generative training strategy} involving generative pre-training and refinement-oriented fine-tuning, to acquire generative priors and improve reconstruction.

%Specifically, we leverage an encoder and a decoder based on the multimodal diffusion image transformer (MMDiT) \cite{esser2024scaling} to facilitate the encoding and reconstruction between the two branches of the 2D image and the 1D encoded sequence.
%Furthermore, we introduce a semantic alignment constraint to directly supervise the encoder in generating compressed tokens that capture accurate high-level semantics aligned with multimodal inputs, thereby enhancing its representational capacity and generalization in downstream tasks.
%Ultimately, we propose an end-to-end two-stage generative training strategy: first, pretraining with a diffusion-based decoder to acquire generative priors and robust representations; then, building a one-step refiner to boost inference efficiency and refine reconstruced textures.

Furthermore, the compression paradigm of the tokenizer dictates the form of AR-based image generation, such as the next-token \cite{ramesh2021zero, yu2022scaling} prediction or next-scale prediction \cite{tian2024visual} with 2D tokenizers.
Building upon SemTok, we explore an AR paradigm tailored for 1D sequences. 
Leveraging their inherent global semantics and close interconnectivity, we construct a masked autoregressive approach to enable high-quality visual generation.
SemTok and the AR model yield distinct improvements in image reconstruction and downstream image generation tasks, demonstrating the potential of semantic 1D tokenization.
%in which we can conduct image generation based on autoregressive modeling, such as the use of next-token prediction \cite{ramesh2021zero, yu2022scaling} or next-scale prediction \cite{tian2024visual} in generative models based on 2D tokenizers.
%Building on our SemTok, we explore the AR paradigm for 1D sequence construction, establishing bidirectional attention across token sequences—characterized by global semantics and tight correlations—via a masked autoregressive modling approach.
Our contributions are as follows:
\begin{enumerate}
\item{
We propose SemTok, a novel visual tokenizer that compresses 2D images into 1D token sequences with compact high-level semantic representations.
Our SemTok achieves state-of-the-art (SOTA) performance on image reconstruction tasks.
}
\item{
Our SemTok improves three key areas: tokenization form, tokenization constraints, and training framework. It validates 1D tokenization and semantic alignment for compact representations, while a generative pipeline enhances the semantic diversity of the latent space.
% Our SemTok explores three key areas: tokenization form, tokenization constraints, and tokenization training framework. It validates the effectiveness of 1D tokenization and semantic alignment constraints for compact representations, while a generative training pipeline is designed to explore a diverse latent space.
}
% \item{
% We introduce a semantic alignment constraint at the encoder to explicitly ensure that encoded tokens possess accurate semantic representations and generalize to multimodal inputs of downstream tasks.
% }
% \item{
% We propose an end-to-end two-stage training strategy that spans diffusion-based pretraining to detokenization-based refinement, aiming to learn high-fidelity image representations and elevate reconstruction quality.
% }
\item{
Building upon SemTok, we construct a masked autoregressive modeling framework for generation with highly competitive performance, which demonstrates the advantages of global 1D semantic tokenization in downstream image generation tasks.
}
\end{enumerate}

\section{Related work}
\paragraph{Image Tokenzation.}
Visual tokenizers project images or videos into compact latent spaces while preserving key visual information.
VAE \cite{kingma2013auto} maps images to continuous distributions, while VQVAE \cite{van2017neural, razavi2019generating} further refines this by quantizing to form discrete vectors.
Subsequent works make improvements in training objectives \cite{ramesh2021zero, esser2021taming, yu2021vector}, multiple quantization \cite{lee2022autoregressive, zheng2022movq}, and quantization techniques \cite{mentzer2023finite, zhao2024image, yu2023language}, aiming to enhance reconstruction quality and facilitate downstream tasks.
However, these works compress images into 2D maps with fixed spatial positions, limiting the scope of semantic representation.

Titok \cite{yu2024image} abandons 2D structure preservation by representing images with learnable 1D tokens, a paradigm extended in works like FlowMo \cite{sargent2025flow}, Flextok \cite{bachmann2025flextok}, and SelfTok \cite{wang2025selftok}.
These works offer new insights into visual tokenizers, yet they may still rely on priors from 2D tokenizers or impose sequential dependencies on tokens.
More critically, these works remain optimized for end-to-end image reconstruction, emphasizing low-level features over high-level semantics, which hinders compact visual representations and their integration with vision-language models (VLMs) \cite{li2023blip, liu2023visual, sun2023emu, DBLP:conf/iclr/Jin0XCLTHCSMZOG24}.
VA-VAE \cite{yao2025reconstruction} and many related works \cite{ma2025unitok,zhao2025qlip} have shown that a pretrained semantic encoder like CLIP \cite{radford2021learning} can achieve concentrated semantic representations and demonstrate strong potential for high-quality generation \cite{zheng2025diffusion}.
In this work, we revisit how to achieve more compact semantic representations and propose an advanced tokenizer by jointly improving the tokenization form, the tokenization constraints, and the training paradigm.

%Abandoning 2D structure preservation, Titok \cite{yu2024image} proposes representing images with learnable 1D tokens, which has been extended in numerous subsequent works (\textit{e.g.}, FlowMo \cite{sargent2025flow}, Flextok \cite{bachmann2025flextok}, and SelfTok \cite{wang2025selftok}).
%However, these works often involve complex training processes, relying on 2D tokenizers to provide priors, or adding constraints for causal dependencies.
%Furthermore, these works optimize for end-to-end image reconstruction tasks, focusing more on restoring low-level features than representing high-level semantics, which is crucial for understanding tasks.
%Many Vision Language Models (VLMs) \cite{li2023blip, liu2023visual, sun2023emu, DBLP:conf/iclr/Jin0XCLTHCSMZOG24} employ CLIP \cite{radford2021learning} encoders to tokenize images with semantic tokens.
%RAE \cite{zheng2025diffusion}, Unitok \cite{ma2025unitok}, VA-VAE \cite{yao2025reconstruction}, and QLIP \cite{zhao2025qlip} indicate that pre-trained semantic encoders can capture semantically rich latents and achieve high-quality generation.
%In this work, we enhance semantic discriminability and consistency by incorporating semantic alignment constraints on top of the 1D image representation.
%We construct a streamlined training pipeline to avoid lengthy training for distilling with other tokenizers.
\begin{figure*}[t]
\centering
\includegraphics[width=0.88\linewidth]{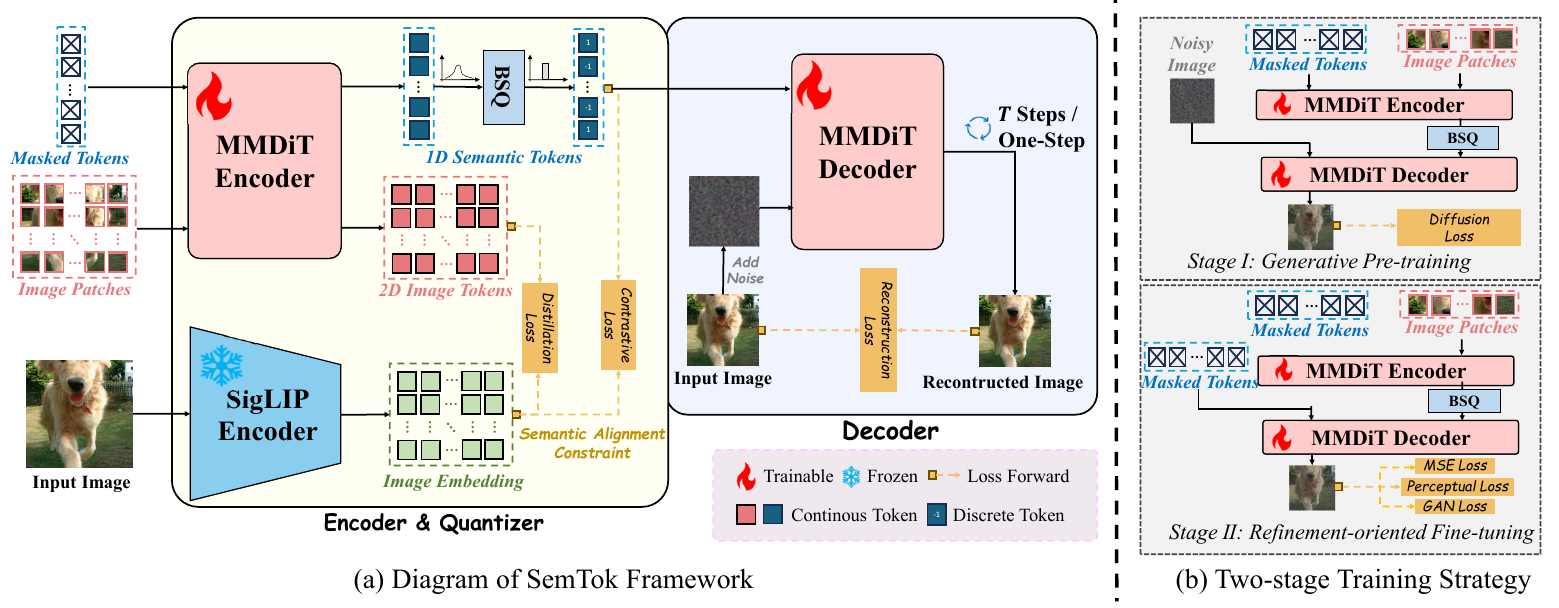}
\caption{An overview of SemTok. (\textit{a}) SemTok compresses 2D images into 1D discrete tokens through the encoder and reconstructs images through the decoder. 
A \textit{semantic alignment constraint} is introduced at the encoder end for compact semantic representations.
(\textit{b}) To explore the semantic diversity of the latent space, SemTok adopts a two-stage training strategy: in Stage I, a diffusion-based decoder is used to predict images from noise; in Stage II,  a one-step refiner is trained on image reconstruction tasks to enhance texture details.}   
\label{fig:tok}
\end{figure*}

\paragraph{Autoregressive Modeling for Generation.}
Diffusion-based image generation \cite{DBLP:conf/nips/DhariwalN21, DBLP:conf/icml/HoogeboomHS23, DBLP:conf/iccv/PeeblesX23, DBLP:conf/iclr/YangTZ00XYHZFYZ25} has achieved remarkable success, while visual autoregressive modeling \cite{tian2024visual, li2024autoregressive, han2025infinity, Yu_2025_ICCV} has recently gained traction, delivering comparable results in many generation tasks. 
These autoregressive methods rely on pre-defined visual tokenizers (e.g., VQVAE \cite{DBLP:conf/iclr/YuLKZPQKXBW22}) to generate images in a step-by-step fashion in 2D sequences. 
We argue that autoregressive generation on 1D tokens is superior compared to 2D tokens, as it captures more compact global semantics for stronger coherence, whereas 2D tokens are constrained by isolated local semantics.
Inspired by bidirectional transformers \cite{DBLP:journals/tmlr/WeberYYDSCC24, zheng2022movq, DBLP:conf/naacl/DevlinCLT19},  we further advocate for masked autoregression, as 1D sequences do not inherently possess nor require the imposition of causal dependencies.
In this paper, building on our SemTok, we explore autoregressive modeling over 1D semantic representations for image generation.

%we believe that the masked autoregression is also preferable, as 1D sequences do not inherently possess nor require the imposition of causal dependencies.
%Image generation methods based on diffusion models \cite{DBLP:conf/nips/DhariwalN21, DBLP:conf/icml/HoogeboomHS23, DBLP:conf/iccv/PeeblesX23, DBLP:conf/iclr/YangTZ00XYHZFYZ25} have achieved great success, and visual autoregressive modeling \cite{tian2024visual, li2024autoregressive, han2025infinity, Yu_2025_ICCV} has garnered more attention recently, delivering comparable results in many generation tasks.
%Autoregressive approaches rely on pre-defined visual tokenizers to generate visual content in a step-by-step fashion (\textit{e.g.}, next-token prediction in 2D sequences based on VQVAE \cite{DBLP:conf/iclr/YuLKZPQKXBW22}).
%Inspired by many bidirectional transformers \cite{DBLP:journals/tmlr/WeberYYDSCC24, zheng2022movq, DBLP:conf/naacl/DevlinCLT19}, we believe that the masked autoregression is also preferable, as 1D sequences do not inherently possess nor require the imposition of causal dependencies.
%In this paper, based on our SemTok, we explore the effectiveness of constructing autoregressive modeling on 1D semantic representations for image generation tasks.

\section{Methods}

We propose SemTok, a novel semantic one-dimensional tokenizer for compact visual representations.
To this end, our SemTok is guided by five core principles that enable effective compression of key visual semantics while ensuring strong scalability across a wide range of downstream tasks.
%Our SemTok adheres to the following four principles to achieve effective compression of key information and ensure its scalability in downstream tasks.
% Our SemTok is a semantic one-dimensional tokenizer for discrete representations. 
% We propose four key principles to adhere to during the design phase, ensuring the effectiveness and scalability of SemTok in image tokenization.
\begin{enumerate}
\item{
\textbf{1D sequence for \textit{global semantics}.} 
SemTok compresses images into 1D representations to avoid redundancy from 2D positional constraints, enabling tokens to capture global semantics and form a compact latent space with strong inter-token correlations.
% SemTok compresses images into 1D representations to avoid the redundancies from 2D tokenizers’ positional constraints. 
% By enabling tokens to capture higher-level global semantics, we achieve a more compact latent space.
% This compact form benefits downstream tasks by fostering strong semantic connections among tokens.
%We aim to compress images into compact 1D representations to avoid the inherent redundancies from 2D tokenizers’ positional constraints, fostering a more compressed latent space.
%This compact form also benefits downstream tasks by forging tight connections between tokens that reflect global semantic information.
}
\item{
\textbf{Semantic representations for \textit{clustered distribution}.} 
%Tokenizers trained on image reconstruction tasks yield low-capacity latents capturing local data intricacies, yet lack the global semantic structure essential for generalization performance \cite{zheng2025diffusion}.
Tokenizers trained on reconstruction yield low-capacity latents rich in local details but poor in global semantics.
Explicit constraints are essential to encourage clustered distributions that encode discriminative information.
%We believe that explicit constraints are essential to encourage clustered latent distributions, thereby encoding discriminative key information.
%Tokenizers trained on image reconstruction produce low-capacity latents that capture local details but lack the global semantics needed for generalization \cite{zheng2025diffusion}.
%We believe that it is essential to impose explicit constraints that encourage the latent space to form clustered distributions, thereby encoding discriminative key information.
% We believe that explicit constraints are essential to encourage clustered latent distributions, thereby encoding discriminative key information.
}
\item{
\textbf{Non-sequential modeling for \textit{bidirectional correlations}.}
Compressed tokens encode complementary semantics with bidirectional dependencies rather than Markovian ones, requiring non-sequential modeling to preserve global correlations without causal constraints.
%Compressed tokens encode global and complementary semantics, exhibiting bidirectional dependencies rather than sequential or Markovian ones.
%without inherently possessing sequential or Markovian dependencies.
%Tokens should be modeled non-sequentially to preserve global correlations without extra causal constraints.
%Tokens should be modeled non-sequentially, without extra causal constraints.
}
\item{
\textbf{Discrete vectors for \textit{semantic disentanglement}.}
Discrete vectors offer clearer semantic disentanglement than continuous latents \cite{wang2025selftok}, enabling easy convergence in the discrete space and allowing us to adopt the LLMs’ successful training paradigm.
%we represent visual data as discrete vectors to adopt LLMs’ successful training paradigm.
%We represent visual data as discrete vectors to inherit and integrate the successful training paradigm of LLMs.
%Additionally, discrete vectors offer clearer semantic disentanglement than continuous latents, enabling easy convergence in the discrete space.
\item{
\textbf{Generative training for \textit{latent space exploration.}} 
%Pixel-level reconstruction may lead to distribution collapse, yielding oversmoothed outputs.
Pixel-level reconstruction may cause distribution collapse and oversmoothed outputs.
Generative training enables the tokenizer to explore a more diverse latent space, allowing sequences to capture richer semantics.
}
}
\end{enumerate}
Based on the principles, in Fig.\ref{fig:tok}, we construct the Semtok framework, which consists of three main components.
(1) An encoder $\mathcal{E}$ compresses the image $I \in \mathbb R^{H\times W\times 3}$ into 1D tokens $z \in \mathbb R^{K\times d}$.
(2) A quantizer $\mathcal{Q}$ looks up the continuous latents $z$ into discrete vectors $\bar{z}$.
(3) A decoder $\mathcal{D}$ generates the reconstructed image $\hat{I}$ given $\bar{z}$.

\subsection{1D Tokenization with Semantic Alignment}
\paragraph{Encoder with \textit{Semantic Alignment Constraint}.} 
Our encoder utilizes an MMDiT-based backbone \cite{esser2024scaling} to achieve the 2D-to-1D conversion, which excels at processing and integrating multimodal inputs (\textit{e.g.}, 2D images and 1D text sequences) through different branches.
We regard 1D tokens as a distinct semantic modality, making 2D-to-1D conversion a natural multimodal alignment task that aligns well with MMDiT’s design.
This dual-stream architecture naturally decomposes 1D tokenization: the 2D branch extracts semantic information, co-attention mechanisms compress it into the 1D branch, and the 1D branch re-integrates the semantics into a structured representation.
We first use the pre-trained VAE of SD3.5 \cite{esser2024scaling} to transform the images $I$ into latents $x_{v} \in \mathbb R^{\frac{H}{f}\times \frac{W}{f}\times D}$ with downscale factor $f=8$, where $x_v$ is then be pachified as the input of one branch.
The other branch inputs a series of learnable embeddings $z_m \in \mathbb R^{K\times d}$ as masked tokens.

%, processing images and tokens in separate branches with co-attention for interaction.
%We believe that this dual-stream architecture can better achieve semantic extraction and compression by splitting from a structured perspective.
%The two branches interact through a co-attention mechanism, where their queries, keys, and values are concatenated.

As discussed earlier, we impose explicit constraints in the encoder to align high-level semantics and ensure a more compact distribution of visual representations.
%We believe it is necessary to incorporate constraints at the encoder for explicit semantic alignment, ensuring visual representations are more compact with high-level semantics. 
Specifically, we introduce a semantic encoder SigLIP \cite{tschannen2025siglip, zhai2023sigmoid} $\mathcal{E}_{sig}$, which has undergone extensive pre-training for language-image alignment, proven effective in understanding tasks, and thus boasts strong cross-modal alignment and semantic extraction capabilities.  
The outputs of our MMDiT encoder and SigLIP encoder are defined as:
\begin{equation}
 \begin{split}
    x, z = \mathcal{E}(x_v, z_m), \ \ x_{sig} = \mathcal{E}_{sig}(I), \\
 \end{split}
\end{equation}
where $x_{sig} \in \mathbb R^{\frac{H}{f}\times \frac{W}{f}\times C}$ is spatially aligned with $x$.
Our \textit{semantic alignment constraint} is divided into two parts, acting on both encoder branches.
In the 2D image branch, we introduce spatial-aligned distillation loss for high-level semantic extraction.
In the 1D token branch, we pool $x_{sig}$ and use contrastive learning loss to associate pairs for the same sample and separate pairs that are not, ensuring the semantics are effectively conversed.
We use quantized results $\mathcal{Q}(z)$ of $z$ herein to compensate for quantization losses.
\begin{equation}
 \begin{split}
 &\mathcal{L}_{\textit{distill}} =  ||x_{sig} - \boldsymbol{w}_x(x)||^2, \\
    \bar{z} = \mathcal{Q}(z),\ &\boldsymbol{u} = \frac{\mathcal{P}(x_{sig})}{||\mathcal{P}(x_{sig})||_2}, \ 
    \boldsymbol{v} = \frac{\boldsymbol{w}_z(\bar{z})}{||\boldsymbol{w}_z(\bar{z})||_2}, \ \\
    \mathcal{L}_{\textit{contra}}&(\boldsymbol{v}, \boldsymbol{u})=\sum_{\mathbf{v} \in \mathcal{V}} (\log \frac{e^{t \boldsymbol{v}^{\top} \boldsymbol{u}}}{\sum_{\mathbf{u} \in \mathcal{U}} e^{t \boldsymbol{v}^{\top} \boldsymbol{u}}}), \\
    &\mathcal{L}_{\textit{sem}} = \mathcal{L}_{\textit{contra}} + \mathcal{L}_{\textit{distill}},\\
 \end{split}
\end{equation}
where $\mathcal{P}$ is the pooling function, $\mathcal{U}, \mathcal{V}$ are collections of samples in a batch,
and $\boldsymbol{w}_x(\cdot), \boldsymbol{w}_z(\cdot)$ are projection weights.
\textit{Semantic alignment constraint} yields strong visual representations, better suited for advanced visual generation tasks.
%\textit{Semantic alignment constraint}  result in strong visual representations, making them more suitable for integration into advanced visual generative tasks.

\paragraph{Binary Spherical Quantizer.}
The quantizer maps continuous latents $z$ to the closest discrete vector $\bar{z}$ in the codebook $\mathcal{C}$, using a straight-through estimator (STE) \cite{bengio2013estimating} to backpropagate through the quantization bottleneck. Increasing the codebook size reduces quantization error but significantly increases memory usage.
Therefore, we employ the Binary Spherical Quantizer (BSQ) \cite{zhao2024image}, akin to Look-up-Free Quantization (LFQ) \cite{yu2023language}, where each channel dimension is mapped to every corner of a hypersphere through binary transformation, with the vector $\bar{z}$ itself also serving as the codebook index $k$.
\begin{equation}
\begin{split}
 \mathcal{Q}(z) = \frac{1}{\sqrt d}sign(\frac{z}{||z||}), \
 k = \sum_{i=0}^{d} \mathbbm{1}(\bar{z}_i>0)\cdot 2^{i-1},
\end{split}
\end{equation}
Therefore, the codebook size of BSQ grows exponentially with channel dimension, with $|\mathcal{C}| = 2^d$. Unlike traditional VQ quantization, BSQ does not require storing a codebook, making it more suitable for scaling up.
%An entropy-based loss is used to promote effective quantization and full codebook utilization \cite{jansen2020coincidence}.
To optimize for effective quantization and encourage codebook utilization, an entropy objective is used as the loss \cite{jansen2020coincidence}:
\begin{equation}
 \begin{split}
 \mathcal{L}_{\textit{quant}} =\mathbb{E}[H(\mathcal{Q}(z))]- H(\mathbb{E}[\mathcal{Q}(z)]).
 \end{split}
\end{equation}
where $H(\cdot)$ represents the entropy function.

\paragraph{Decoder.} 
Our decoder also utilizes an MMDiT-based architecture. 
Since image reconstruction is a generative task for the decoder and requires strong generative priors to capture a richer semantic spectrum \cite{wang2025selftok}, we employ a decoder with larger parameters than the encoder and initialize its parameters from SD3.5 \cite{esser2024scaling}.
We apply Relative Positional Encoding (RoPE) \cite{su2024roformer} to better represent positional relationships.
%Since image reconstruction is a generative task for the decoder, we equip the decoder with a stronger generative prior to capture a richer semantic spectrum, initializing its parameters from SD3.5 \cite{esser2024scaling}.
%Since image reconstruction is a generative task for the decoder, following SelfTok \cite{wang2025selftok}, we initialize the decoder's parameters with SD3.5 \cite{esser2024scaling}.
%The decoder is the reverse process of the encoder, where the semantics are transferred from the 1D token branch to another branch, generating a 2D image from noise or masked tokens.

\subsection{Two-stage Training for Image Reconstruction}
To explore a richer semantic distribution in the latent space, we design a two-stage generative training strategy, where the decoder generates images from noise or masked tokens.
%with stage-specific objectives.
\paragraph{Generative Pre-training.}
Pixel-level reconstruction constraints, aimed at optimizing fidelity, tend to converge to an overly smoothed latent space, making it difficult to generate semantically rich and realistic textures \cite{karras2019style}.
To better explore the latent space, we first employ diffusion-based generative pre-training, as it optimizes the likelihood $p(\hat{I}|z)$ across all noise scales, preserving multiple distribution paths instead of collapsing to a single solution \cite{rombach2022high}.
Diffusion-based decoders, leveraging the principled probabilistic modeling of the Evidence Lower Bound (ELBO), can more effectively capture the richness and diversity of semantic patterns \cite{chen2025diffusion}.
Building on flow matching \cite{DBLP:conf/iclr/LiuG023}, the decoder models the velocity distribution from noisy image latents $x_t$, with noise $\epsilon$ and time $t \in [0,1]$.
The objective is as follows:
%In the first stage, our objective is to jointly train the encoder and decoder, enabling the encoded tokens $z$ to maximize the extraction and representation of high-level semantics while equipping the decoder with generative priors to achieve restoration of semantic content.
%Given that diffusion models can model and optimize likelihood distributions $p(\hat{I}|z)$ from noise, they thus circumvent the limitation of over-fidelity optimization of pixel-level constraint \cite{karras2019style}, which may struggle to generate rich textures.
%Building on flow matching \cite{DBLP:conf/iclr/LiuG023}, the decoder models the velocity distribution from noisy image latents $x_t$, with noise $\epsilon$ and time $t \in [0,1]$. The objective is as:
\begin{equation}
 \begin{split}
 &x_t = t\epsilon +(1-t)x_v,\\
 \mathcal{L}_{\textit{diff}} = &\mathbb{E} \Big [||x_v-\epsilon - \mathcal{D}(x_t, \bar{z}, t)||^2 \Big],
 \end{split}
\end{equation}
In the generative training, we jointly optimize the encoder $\mathcal{E}$, quantizer $\mathcal{Q}$, and decoder $\mathcal{D}$ with the objective as:
\begin{equation}
 \begin{split}
 \min \ \  \mathcal{L}_{\textit{diff}} + \lambda_q \mathcal{L}_{\textit{quant}} + \lambda_s \mathcal{L}_{\textit{sem}},  
 \end{split}
\end{equation}
Flow-matching pre-training yields a more diverse latent space as a prior for image reconstruction, enabling encoded tokens $z$ to maximally represent high-level semantics.
%During this phase, we prioritize learning semantically rich tokenization to mitigate the impact of over-fidelity optimization on semantic representation.
%The diffusion-based decoder is applicable for producing high-fidelity images via multi-step denoising.

\paragraph{Refinement-oriented Fine-tuning.}  
Although the first-stage pretraining yields a rich latent space, its fidelity objective is optimized in the latent space rather than the image space, ignoring the compression loss incurred during the pixel-to-latent encoding. 
Such manipulation degrades the generation of fine textures, where the inherent limitations of VQVAE restrict the tokenizer's capacity ceiling.
Moreover, the diffusion-based decoder incurs a high inference cost due to iterative sampling.
%Additionally, the diffusion-based decoder necessitates multiple iterations during inference, leading to additional computational overhead.
Therefore, we conduct refinement-oriented training to achieve superior image reconstruction.
%As the optimization objective for fidelity in the first stage is performed in the latent space rather than the image space, the compression loss incurred from pixel-to-latent conversion is overlooked, which may compromise the generation of detailed textures.
%The inherent limitations of VQVAE restrict the tokenizer's capacity ceiling.

In the decoder, the input of the original 2D noise branch is replaced by a sequence of learnable masked tokens $z_m$, directly reconstructing the image latent.
The original diffusion $v$-prediction loss in the optimization objective is replaced with pixel-level reconstruction loss, which includes the MSE, perceptual, and GAN losses \cite{goodfellow2020generative}, enabling SemTok to address potential detail losses resulting from latent compression adaptively.
%By providing pixel-level contraint, SemTok can adaptively address potential detail losses resulting from latent compression.
%We freeze the encoder and quantizer, as the semantic alignment achieved in the first phase has secured rich semantic representations. Only the decoder is fine-tuned, with the input to the 2D noisy image branch replaced by a sequence of learnable mask tokens $z_m$.
%Our objectives include the classic MSE loss for pixel-level reconstruction, perceptual loss, and GAN loss \cite{goodfellow2020generative}.
\begin{equation}
 \begin{split}
 &\hat{I} = \mathcal{D}(\bar{z}, z_m),\\
 \min \ \mathcal{L}_{\textit{rec}} = ||I-\hat{I}||^2 &+ \lambda_p \mathcal{L}_{\textit{per}}(I, \hat{I}) + \lambda_G \mathcal{L}_{\textit{G}}(I,\hat{I}).
 \end{split}
\end{equation}
Following \cite{lipman2022flow} and \cite{DBLP:journals/tmlr/WeberYYDSCC24}, besides the classical LPIPS \cite{zhang2018unreasonable}, we incorporate ResNet \cite{he2016deep} loss into the perceptual loss to further enhance reconstruction performance.

\subsection{Autoregression for Image Generation}
\begin{figure}[t]
\centering
\includegraphics[width=0.9\linewidth]{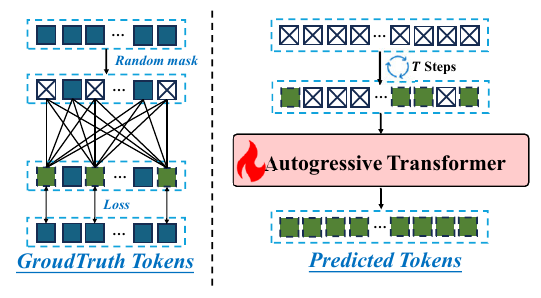}
\caption{Illustration of masked autoregressive modeling of image generation with SemTok. (\textit{left}) During training, we randomly mask several tokens and utilize bidirectional attention to allow each token to see all tokens, thereby perceiving global semantics. (\textit{right}) During inference, the model predicts multiple tokens simultaneously in a random order until the complete sequence is filled.}   
\label{fig:ar}
\end{figure}
A proficient tokenizer benefits downstream generative tasks via exceptional image representation.
Leveraging SemTok, we build an AR framework to validate our semantic 1D tokenization for image generation.
%SemTok compresses images into a compact token sequence that captures global semantics with stronger mutual dependencies.
SemTok compresses images into a compact token sequence that serves as constituents of global semantics, exhibiting stronger mutual dependencies. 
This enables greater gains in autoregressive generation, as previously generated tokens provide a stronger prior for subsequent ones.
Since our 1D tokenization models bidirectional dependencies between tokens rather than sequential Markovian ones, it enables parallel token prediction with global context, thereby expanding the inherently unidirectional receptive scope of next-token prediction.
Following MaskGIT \cite{chang2022maskgit}, MAR \cite{li2024autoregressive}, and Muse \cite{DBLP:conf/icml/ChangZBML00MFRL23}, we adopt bidirectional transformers for masked autoregressive modeling.
%Since our 1D tokenization imposes no causal constraints and semantic tokens lack inherent sequentiality, they cannot be generated via Markovian next-token prediction.
%Following \cite{chang2022maskgit, li2024autoregressive,DBLP:conf/icml/ChangZBML00MFRL23}, we instead adopt bidirectional masked transformers for parallel token generation.

%Without causal dependency constraints on tokens during 1D tokenization, these tokens cannot be generated via the next-token prediction form that satisfies the Markov property.
%Thus, we follow the format of works such as MaskGIT \cite{chang2022maskgit}, MAR \cite{li2024autoregressive}, and Muse \cite{DBLP:conf/icml/ChangZBML00MFRL23}, generating tokens through bidirectional masked transformers.
%We argue that these compact representations, as integral components of global semantics and closely interrelated, can better acquire priors via the autoregressive process.

As shown in Fig.~\ref{fig:ar}, during training, we mask tokens in $\bar{z}$ randomly with a binary mask $M$ to form a masked token sequence $z_m$.
%random tokens in $\bar{z}$ are replaced with a masked token $z_m$. 
The transformer $\theta$ with bidirectional attention enables mutual perception between masked and known tokens, predicting the distribution probabilities of masked tokens.
During inference, at the $n$-th step across $N$ iterative steps, the transformer $\theta$ predicts the masked tokens $\hat{z}^n$ at several random positions based on the current known tokens $(\hat{z}^1, \ldots, \hat{z}^{n-1})$ and binary mask $M_n$. 
The framework updates the token sequence and binary mask, gradually transitioning from a sequence of masked tokens to semantic tokens, accelerating inference by generating multiple tokens simultaneously.
The process is defined as follows: 
\begin{equation}
 \begin{split}
 &\hat{z}^n = (M_{n-1} \oplus M_n) \circ \theta(\hat{z}^1, \ldots, \hat{z}^{n-1}), \\
 p&(\hat{z}^1, \ldots, \hat{z}^N)=\prod_n^N p\left(\hat{z}^n \mid \hat{z}^1, \ldots, \hat{z}^{n-1}\right).
 \end{split}
\end{equation}
%For class-conditioned image generation, we concatenate the learnable class token with the image tokens.
%and place it at the beginning. Additionally, it serves as a condition through AdaLN \cite{DBLP:conf/iccv/PeeblesX23}.
%Our autoregressive model follows the architecture of LlamaGen \cite{sun2024autoregressive} with 1D RoPE \cite{su2024roformer}.

\section{Experimental Results}
\subsection{Implementation Details}
\paragraph{Image Recontruction}
We train and evaluate tokenizers on ImageNet-1k \cite{deng2009imagenet} at 256×256 resolution using standard metrics: rFID \cite{heusel2017gans}, PSNR, SSIM, and LPIPS \cite{zhang2018unreasonable}.
We provide the codebook size $|\mathcal{C}|$, token length $K$, and Bits-per-pixel as the compression ratio $\frac{log_2(|\mathcal{C}| )\cdot K}{H W}$.
SemTok employs FSQ \cite{zhao2024image} for quantization, differing from VQ \cite{van2017neural} by not needing extra codebook storage.
We provide two models with different codebook sizes to evaluate their impact on the reconstruction quality.

% We evaluate the performance of different tokenizers on image reconstruction tasks after training on ImageNet-1k \cite{deng2009imagenet} with resolution of $256\times256$ (except SD-VAE \cite{rombach2022high}).
% We use widely adopted metrics for evaluation of the reconstructions, including rFID \cite{heusel2017gans}, PSNR, SSIM, and LPIPS \cite{zhang2018unreasonable}.
%We provide the codebook size $|\mathcal{C}|$, token length $K$, and Bits-per-pixel as the compression ratio $\frac{log_2(|\mathcal{C}| )\cdot K}{H W}$.
%Our SemTok uses FSQ \cite{zhao2024image} for quantization, unlike VQ \cite{van2017neural}, no extra codebook storage is needed.
%We thus train two models with different codebook sizes to verify the correlation between this growth and reconstruction quality.

%Our encoder and decoder follows the MMDiT with the decoder initialized from SD3.5 \cite{esser2024scaling}.
Images are compressed using SD3.5-VAE \cite{rombach2022high} with a downscale factor of 8 into quantized tokens with length $256$.
The hyperparameters $\lambda_q, \lambda_s, \lambda_p,\lambda_G$ are set to $0.0025, 0.05, 0.1, 0.1$, respectively.
We train SemTok for 200k iterations in Stage \uppercase\expandafter{\romannumeral1} with batchsize 256 and 100k iterations in Stage \uppercase\expandafter{\romannumeral2} with batchsize 64, using AdamW with learning rate
of $5\times 10^{-5}$ and weight decay of $1\times 10^{-3}$.

%We train SemTok for 200k iterations in Stage \uppercase\expandafter{\romannumeral1} with batchsize 128 and 50k iteration in Stage \uppercase\expandafter{\romannumeral2} with batchsize 64.
%We employ the AdamW with learning rateof $5\times 10^{-5}$ and weight decay of $1\times 10^{-3}$.
%We use the SD3.5-VAE \cite{rombach2022high} to compress images with downscale factor 8.
%The quantized code length is 256.
%The encoder's structure is based on \cite{sargent2025flow}, with a depth of 8 layers, hidden layer dimension of 1152 and patch size of 2.
% We use the exponential moving average (EMA) with rate 0.999.
% All experiments are conducted on 8 NVIDIA H20 GPUs.

\begin{table*}[t]
    \scriptsize
  \centering
  \caption{Reconstruction results of SOTA tokenizers on ImageNet $(256\times256)$. *:  indicates a continuous tokenizer trained on OpenImages. -re: indicates the use of rejection sampling.}
  \label{tab:recon_result}
  \begin{tabular}{c|c|cccc|cccc}
    \toprule
    Tokenizer & Type & Quant & $\#$Bits-per-pixel&  $\#$Token& $\#$Codebook & rFID$\downarrow$ &  PSNR $\uparrow$ & SSIM$\uparrow$ & LPIPS$\downarrow$ \\ 
    \midrule
Taming VQ-GAN-re \cite{esser2021taming} &2D& VQ &0.055 & 16×16& $2^{14}$ & 4.98&- &- &-\\

LlamaGen-16 \cite{sun2024autoregressive} &2D&VQ&0.055 & 16×16& $2^{14}$ &2.19 &20.67 &0.589 &0.132\\
MaskBiT \cite{DBLP:journals/tmlr/WeberYYDSCC24} &2D&LFQ&0.055& 16×16& $2^{14}$ &1.37 &21.50& 0.560&-\\
 Cosmos DI-16x16 \cite{agarwal2025cosmos} &2D&FSQ &0.062&16×16  &$\approx2^{16}$ &4.40& 19.98& 0.536& 0.153\\
OpenMagViT-V2 \cite{luo2024open} &2D&LFQ &0.070 & 16×16 & $2^{18}$ &1.17& 21.63& 0.640& 0.111\\
VAR \cite{tian2024visual} &2D & VQ&0.125 & 680& $2^{12}$& 0.99 &22.12 &0.624 &0.109\\
RQ-VAE \cite{lee2022autoregressive} &2D& VQ &0.055 & 8×8×4 & $2^{14}$& 3.20& - &- &-\\
ViT-VQGAN \cite{yu2021vector} &2D & VQ &0.203&32×32& $2^{13}$& 1.28& - &- &-\\
SD-VAE$^{*}$ \cite{rombach2022high} &2D &KL& 1.000 &32*32 & - &1.35 &21.99 & 0.628 & \textbf{0.098}\\
    \midrule
    TiTok-L-32 \cite{yu2024image}&1D&VQ&0.006 &32 &$2^{12}$ & 2.21 &15.60& 0.359& 0.322\\
    TiTok-B-64\cite{yu2024image}& 1D&VQ &0.012&64 &$2^{12}$& 1.70& 16.80 &0.407& 0.252\\
    TiTok-S-128 \cite{yu2024image}& 1D&VQ&0.023&128 &$2^{12}$&1.71 &17.52& 0.437 &0.210\\
    FlexTok \cite{bachmann2025flextok}& 1D &FSQ&0.062&256& $64,000$&1.45& 18.53& 0.465& 0.222\\
    FlowMo-Lo \cite{sargent2025flow}& 1D &LFQ &0.070 &256& $2^{18}$ &0.95& 22.07& 0.649& 0.113\\
    \midrule
\textbf{SemTok (Ours)} & 1D &BSQ &0.070 &256& $2^{18}$ & 0.88 & 22.19& 0.639 &0.128\\
\textbf{SemTok (Ours)} & 1D &BSQ &0.125&256& $2^{32}$ & \textbf{0.67}& \textbf{23.05}& \textbf{0.684}& 0.100\\
    \bottomrule
  \end{tabular}
\end{table*} 
\begin{figure*}[t]
\centering
\includegraphics[width=0.84\linewidth]{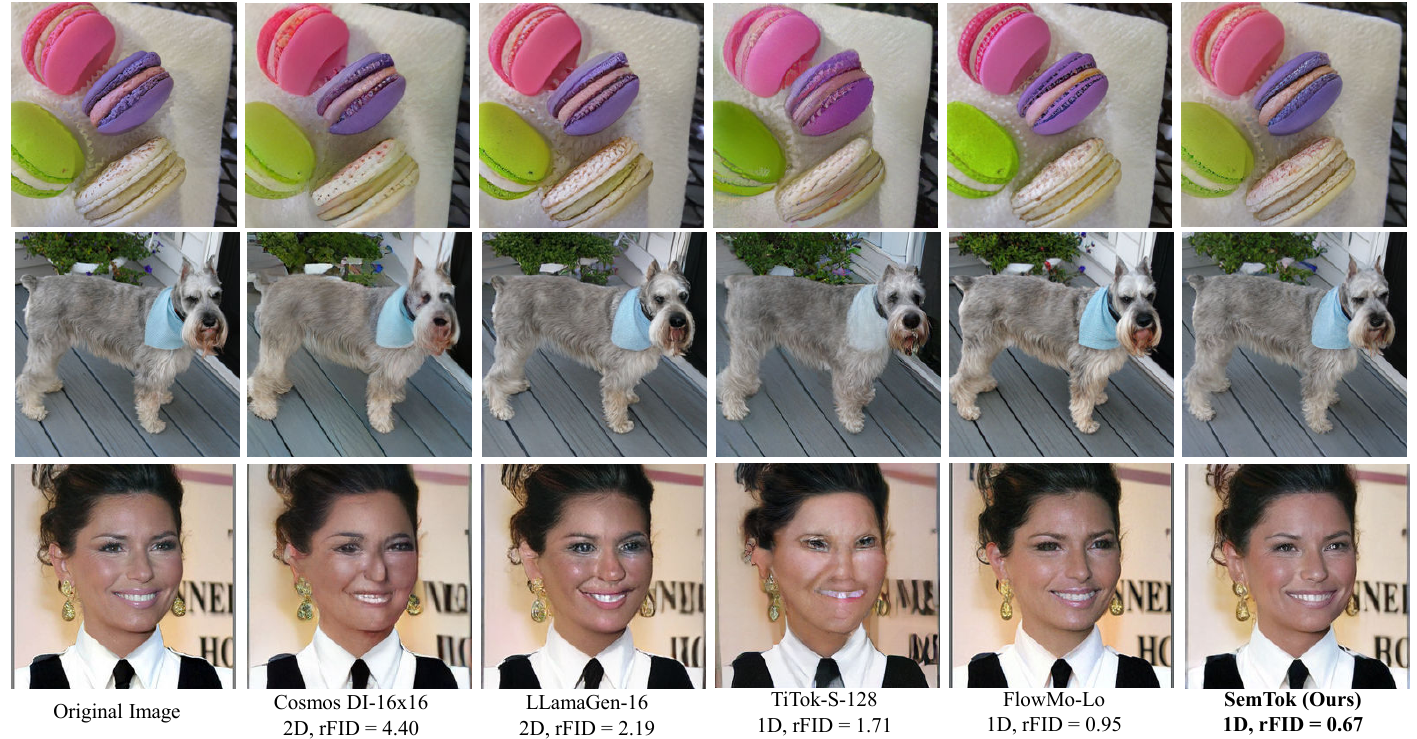}
\caption{Comparison of reconstructions from different tokenizers.}   
\label{fig:recon}
\end{figure*}
\begin{figure*}[t]
\centering
\includegraphics[width=\linewidth]{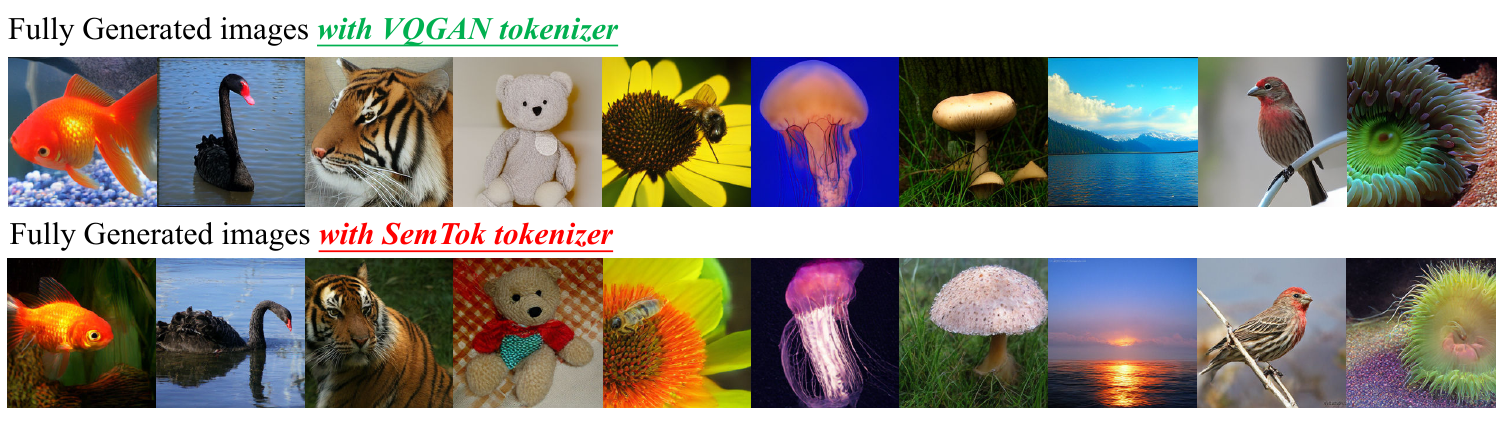}
\vspace{-5mm}
\caption{Generation results with different tokenizers.}   
 \vspace{-3mm}
\label{fig:generation_tokenizer}
\end{figure*}

\paragraph{Image Generation.}
We build a masked autoregressive framework based on SemTok and evaluate it on class-conditional image generation using ImageNet-1k (256×256) \cite{deng2009imagenet}.
We report standard metrics: gFID \cite{heusel2017gans}, Inception Score (IS) \cite{salimans2016improved}, Precision, and Recall \cite{kynkaanniemi2019improved}.

%We construct the masked autoregressive framework based on our SemTok and validate it on the task of class-to-image generation task on ImageNet-1k ($256\times 256$) \cite{deng2009imagenet}.
%We use widely adopted metrics to measure the reconstructions, including gFID \cite{heusel2017gans}, Inception Score (IS) \cite{salimans2016improved}, Precision, and Recall \cite{kynkaanniemi2019improved}.

AR model adopts the LLamaGen \cite{sun2024autoregressive} with bidirectional attention.
It is trained for 800k iterations with batchsize 2048, using AdamW with learning rate of $2\times 10^{-4}$ and weight decay of $0.03$.
During training, we randomly sample a ratio from [0.7, 1.0] to mask semantic tokens \cite{he2022masked, li2024autoregressive}. 
At inference, tokens are filled in random masked positions over 64 steps via cosine annealing, with Classifier-Free Guidance (CFG) \cite{ho2022classifier} applied at a linearly increased scale of 6.0.
% During inference, the AR model gradually fills all masked tokens in the sequence at random positions using cosine annealing, with 20 steps by default. 
% We apply Classifier-Free Guidance (CFG) \cite{ho2022classifier} with a constant scale of 5.5 at each iteration step.

%We apply a dropout probability of 0.1 on the class condition to enable Classifier-Free Guidance (CFG) \cite{ho2022classifier}. 

\subsection{Image Recontruction}
\textbf{Quantitative results.} 
Quantitative results for image reconstruction are shown in Tab.~\ref{tab:recon_result}. Compared to conventional 2D and 1D tokenizers, SemTok achieves superior rFID and PSNR at the same Bits-per-pixel of $0.070$.
This gain stems from SemTok’s unique design: by enforcing \textit{semantic alignment constraint} and representing images as compact 1D high-level tokens, it captures global structure more effectively while avoiding the redundancy and local bias of pixel-aligned 2D grids.
By further increasing the codebook size to expand representational capacity without adding storage, SemTok achieves consistent improvements and \textbf{attains SOTA results across most metrics with an impressive $0.67$ rFID on ImageNet}, even surpassing the continuous VAE tokenizer of SD-v2.1 \cite{rombach2022high}.

\textbf{Qualitative results.} 
In Fig.~\ref{fig:recon}, we visualize some reconstruction results between different tokenizers from the test set.
Our SemTok consistently generates images with accurate semantics and fine-grained texture details, aligning with the metrics in Tab. \ref{tab:recon_result}.
For example, SemTok demonstrates superior structural fidelity over other tokenizers, excelling in creating intricate animal fur, facial features, and detailed structures.
The results highlight SemTok’s strong capability to compress accurate semantic representations and generate realistic images, even under a compact compression ratio.

\subsection{Image Generation}
\begin{table}[t]
    \footnotesize
  \centering
  \caption{Generation results with other methods on ImageNet ($256\times256$).  *: indicates generating images at resolution $384\times 384$ and resizing back to $256\times256$. }
  \label{tab:generation_result}
  \resizebox{1.0\linewidth}{!}{
  \begin{tabular}{c|cc|cccc}
    \toprule
    Generator & Type & Params & gFID$\downarrow$ &  IS $\uparrow$ & Pre$\uparrow$ & Rec$\uparrow$ \\ 
    \midrule
BigGAN \cite{brock2018large} & GAN &112M& 6.95 &224.5 & 0.89& 0.38\\ 
GigaGAN \cite{kang2023scaling}& GAN& 569M& 3.45 &225.5& 0.84& 0.61\\
\midrule
 LDM-4-G \cite{rombach2022high}&Diff.&400M &3.60& 247.7& - &- \\
DiT-L/2 \cite{DBLP:conf/iccv/PeeblesX23}& Diff. &458M& 5.02 &167.2& 0.75& 0.57\\
DiT-XL/2 \cite{DBLP:conf/iccv/PeeblesX23}&Diff. &675M & 2.27 &278.2& 0.83& 0.57\\
VDM++ \cite{DBLP:conf/nips/KingmaG23}&Diff.& 2.0B &2.12& 267.7&-&-\\ 
\midrule
VQGAN \cite{esser2021taming}&AR&1.4B& 15.78& 78.3&-&-\\
RQ-Tran. \cite{lee2022autoregressive} &  AR &3.8B &3.80 &323.7 &- &-\\
ViTVQ \cite{yu2021vector}&AR&1.7B& 4.17& 175.1& - &- \\ 
LlamaGen-L$^{*}$ \cite{sun2024autoregressive}&AR& 343M&3.07& 256.1 &0.83 &0.52\\
LlamaGen-XL$^{*}$ \cite{sun2024autoregressive}&AR&775M& 2.62& 244.1& 0.80 &0.57\\
LlamaGen-XXL$^{*}$ \cite{sun2024autoregressive}&AR&1.4B& 2.34& 253.9 &0.80& 0.59\\
LlamaGen-3B$^{*}$ \cite{sun2024autoregressive}&AR&3.1B& 2.18& 263.3& 0.81& 0.58\\
RandAR-L \cite{pang2025randar}&AR&343M& 2.55& 288.8& 0.81 &0.58 \\
RandAR-XL \cite{pang2025randar}&AR&775M& 2.25& 317.8& 0.80 &0.60\\
RandAR-XXL\cite{pang2025randar}&AR&1.4B& 2.15& 322.0& 0.79& 0.62\\
Open-MAGVIT2-B \cite{luo2024open}&AR& 343M &3.08 &258.3 &0.85 &0.51\\
Open-MAGVIT2-L \cite{luo2024open}&AR& 804M &2.51& 271.7 &0.84 &0.54\\
Open-MAGVIT2-XL \cite{luo2024open}&AR& 1.5B &2.33& 271.8 &0.84 &0.54\\
VAR-d16 \cite{tian2024visual} &AR & 310M &3.30& 274.4& 0.84& 0.51\\
VAR-d20 \cite{tian2024visual} &AR & 600M &2.57 &302.6& 0.83& 0.56\\
VAR-d24  \cite{tian2024visual} &AR&1.0B& 2.09& 312.9& 0.82& 0.59\\ 
MAGE \cite{li2023mage} & AR & 230M& 6.93& 195.8 &-&-\\
MaskGIT \cite{chang2022maskgit} &AR &227M& 6.18 &182.1 &0.80 &0.51\\
FlowMo-Lo \cite{sargent2025flow} & AR&397M&4.30& 274.0 & 0.86& 0.47\\
\midrule
\textbf{SemTok-AR-L (Ours)} & AR& 318M &2.77& 293.1 & 0.78& 0.60\\
\textbf{SemTok-AR-XL (Ours)} & AR& 746M & 2.54 & 305.6 & 0.78 & 0.60\\
\textbf{SemTok-AR-XXL (Ours)} & AR& 1.2B &2.34 & 310.5 & 0.77 & 0.63\\
\bottomrule
  \end{tabular}}
 \vspace{-3mm}
\end{table} 
Based on SemTok, we build a masked autoregressive framework for image generation and compare it against previous GAN-based, diffusion-based, or AR-based methods, with results shown in Tab.~\ref{tab:generation_result}.
We further explore different sizes of the AR model (from 318M to 1.2B) and observe strong scalability with consistent performance improvements.
\textbf{Our AR model achieves performance on generation tasks fully comparable to SOTA methods} VAR \cite{tian2024visual}, LlamaGen \cite{sun2024autoregressive}, and RandAR \cite{pang2025randar}.
Notably, compared to MaskGIT \cite{chang2022maskgit} and MAGE \cite{li2023mage}, which are also based on masked autoregressive modeling with classical 2D tokenizers, our SemTok-based approach significantly advances the generation quality of this paradigm.
Those results validate the scalability of 1D semantic representations for downstream tasks, enabling tighter global modeling in the autoregressive process through more compact semantic tokens.

\subsection{Ablation Study}
\begin{table}[t]
    \scriptsize
  \centering
  \caption{Generation results of different tokenizers on ImageNet ($256\times256$).}
  \label{tab:generation_tok}
  \begin{tabular}{c|c|cccc}
    \toprule
    Tokenizer & Type & gFID$\downarrow$ &  IS $\uparrow$ & Pre$\uparrow$ & Rec$\uparrow$ \\ 
    \midrule
VQGAN \cite{esser2021taming}&2D& 4.62 & 254.1 & \textbf{0.84}& 0.49\\
FlowMo-Lo \cite{sargent2025flow} & 1D& 6.68 & 280.4 & 0.82 & 0.45\\
\midrule
\textbf{SemTok (Ours)} & 1D& \textbf{2.77}& \textbf{293.1} & 0.78& \textbf{0.60}\\
\bottomrule
  \end{tabular}
\end{table} 
\begin{figure}[t]
\centering
\includegraphics[width=\linewidth]{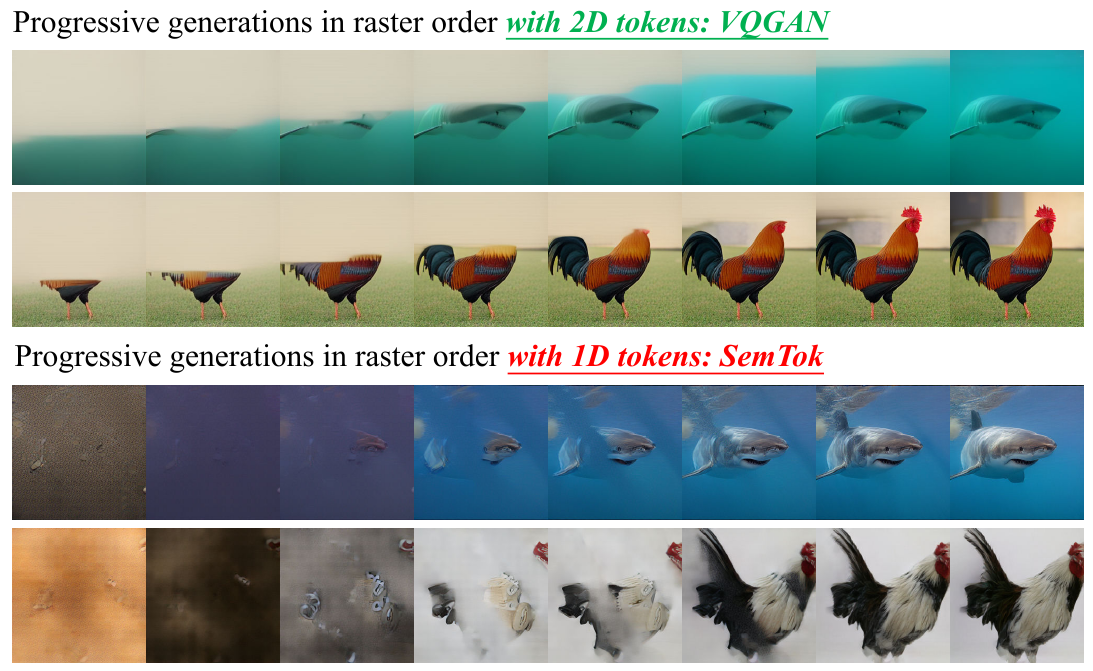}
\vspace{-3mm}
\caption{Progressive generation results in raster scan order between 1D and 2D tokens.}   
\vspace{-3mm}
\label{fig:iterative}
\end{figure}
\paragraph{Generation with different tokenizers.}
In Tab.~\ref{tab:generation_tok}, we compare SemTok against two baseline tokenizers on image generation using the same AR framework.
\textbf{SemTok significantly outperforms both 2D grid-based VQGAN \cite{esser2021taming} and previous 1D-based FlowMo-Lo \cite{sargent2025flow}}.
The superiority of SemTok aligns with our view that 1D representations capture global semantics unbound by spatial structure, enabling richer and more coherent autoregressive generation.
Further boosted by \textit{semantic alignment constraint} and \textit{generative training strategy}, SemTok learns more compact, high-level visual representations, yielding better downstream performance.

\begin{figure}[t]
\centering
\includegraphics[width=0.9\linewidth]{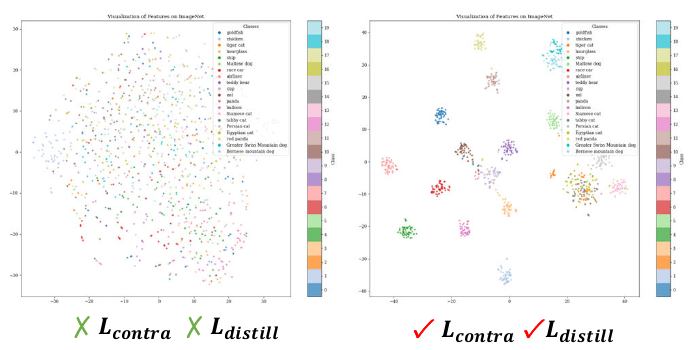}
\caption{Feature Clustering with different loss functions.}   
\vspace{-3mm}
\label{fig:generation}
\end{figure}

We visualize generated samples based on VQGAN and SemTok tokenizers in Fig.~\ref{fig:generation}, demonstrating that SemTok produces high-quality images with superior fidelity and diversity.
To further highlight the semantic distinction between 1D and 2D tokens, Fig.~\ref{fig:iterative} visualizes multi-step autoregressive generation using raster-order decoding.
2D tokens follow a rigid spatial pattern, with each token encoding only local content, while 1D tokens progressively refine global semantics without spatial constraints.
This aligns with our view that 1D semantic encoding compresses global semantics, yielding more compact and semantic-preserved representations.

%Moreover, to better illustrate the semantic differences between 1D and 2D tokens, we visualize their autoregressive generation processes over multiple steps in Fig.~~\ref{fig:iterative}, using raster-order decoding for both.
%The 2D tokens exhibit a clear spatially ordered generation pattern, with each token encoding only local semantics within its corresponding image region.
%In contrast, our 1D tokens clearly break free from spatial constraints, progressively enriching and refining global semantics as more tokens are generated.
%This aligns with our view that 1D semantic encoding compresses global semantics, yielding more compact and semantically faithful representations.

\begin{table}[t]
\caption{Ablation on the \textit{Semantic Alignment Constraint}.}
\label{tab:feature}
\scriptsize
\centering
\begin{tabular}{ cc | cc |cc }
\toprule
\multicolumn{2}{c|}{Constraint} & \multicolumn{2}{c|}{Reconstruction} &  \multicolumn{2}{c}{Generation}\\
%\cmidrule(lr){2-11}
Contrastive & Distill & rFID$\downarrow$  & PSNR$\uparrow$ & gFID$\downarrow$ &  IS $\uparrow$ \\
\midrule
\XSolidBrush & \XSolidBrush  & 1.08& 21.74& 3.83 & 271.4\\
\Checkmark & \XSolidBrush  & 0.97 & 21.86& 2.87 & 276.8\\
\XSolidBrush & \Checkmark  & 1.02 & 21.79& 3.55 & 264.9\\
\midrule
\Checkmark& \Checkmark & \textbf{0.88}& \textbf{22.19}& \textbf{2.77}& \textbf{293.1} \\
\bottomrule
\end{tabular}
\end{table}
\paragraph{Semantic Alignment Constraint.}
In Tab.~\ref{tab:feature}, we compare the results of applying versus not applying the \textit{semantic alignment constraint} on dual branches of the encoder.
The constraints on both branches consistently improve both reconstruction and downstream generation performance.
Furthermore, we visualize the clustering of encoder features (\textit{i.e.}, the layer immediately preceding token compression) in the embedding space under different settings. 
As shown, applying the \textit{semantic alignment constraint} leads to clear clustering of features in the embedding space, indicating that key semantic concepts are explicitly preserved and emphasized.
By aligning learned tokens with high-level semantic features from a pretrained vision model, our constraint guides the tokenizer to preserve meaningful visual concepts over low-level artifacts, enhancing semantic fidelity and bridging discrete tokenization with high-level visual understanding.

\begin{table}[t]
\caption{Ablation on the \textit{Masked Autoregressive Modeling} vs. \textit{Next-token Prediction}.}
\label{tab:order}
\scriptsize
\centering
\begin{tabular}{ c | cc |cc }
\toprule
\multirow{2}{*}{Order} & \multicolumn{2}{c|}{Reconstruction} &  \multicolumn{2}{c}{Generation}\\
%\cmidrule(lr){2-11}
~ & rFID$\downarrow$  & PSNR$\uparrow$ & gFID$\downarrow$ &  IS $\uparrow$ \\
\midrule
Raster & 1.22& 21.07 & 3.08 & 269.9\\
Global & \textbf{0.88}& \textbf{22.19} &\textbf{2.77}& \textbf{293.1}\\
\bottomrule
\end{tabular}
\end{table}
\paragraph{Non-sequential modeling.}
Our SemTok adopts a non-sequential modeling approach and employs a masked autoregressive framework for image generation.
In Tab.~\ref{tab:order}, we compare against the sequential 1D tokenization in SelfTok \cite{wang2025selftok}, which generates tokens autoregressively via next-token prediction.
Non-sequential modeling achieves superior results in both image reconstruction and generation, as it respects the inherent global semantic complementarity among tokens, enables more effective modeling of global correlations, and eliminates the inductive bias introduced by arbitrary token ordering.
We argue that imposing additional sequential constraints compromises the information density of the compressed sequence and hinders its ability to model long-range dependencies.

\begin{table}[t]
    \begin{minipage}[t]{\linewidth}
  \centering
        \scriptsize
        \caption{Ablation on the \textit{Two-stage Training Strategy}.}
      \label{tab:refiner}
       \centering
       \begin{tabular}{ c |c | cccc }
\toprule
Tokenizer & $\#$Codebook & rFID$\downarrow$ &  PSNR $\uparrow$ & SSIM$\uparrow$ & LPIPS$\downarrow$\\
\midrule
Diffusion& $2^{18}$& 1.31 & 20.48 & 0.580& 0.177\\
Refiner&$2^{18}$& \textbf{0.88} & \textbf{22.19}& \textbf{0.639} &\textbf{0.128}\\
\midrule
Diffusion& $2^{32}$& 1.03 & 21.43 & 0.633& 0.145\\
Refiner& $2^{32}$&\textbf{0.67}& \textbf{23.05}& \textbf{0.684}& \textbf{0.100}\\
\bottomrule
\end{tabular}
    \end{minipage}\\
    \vskip 0.15in
    \begin{minipage}[t]{\linewidth}
		  \centering
        \includegraphics[width=0.9\linewidth]{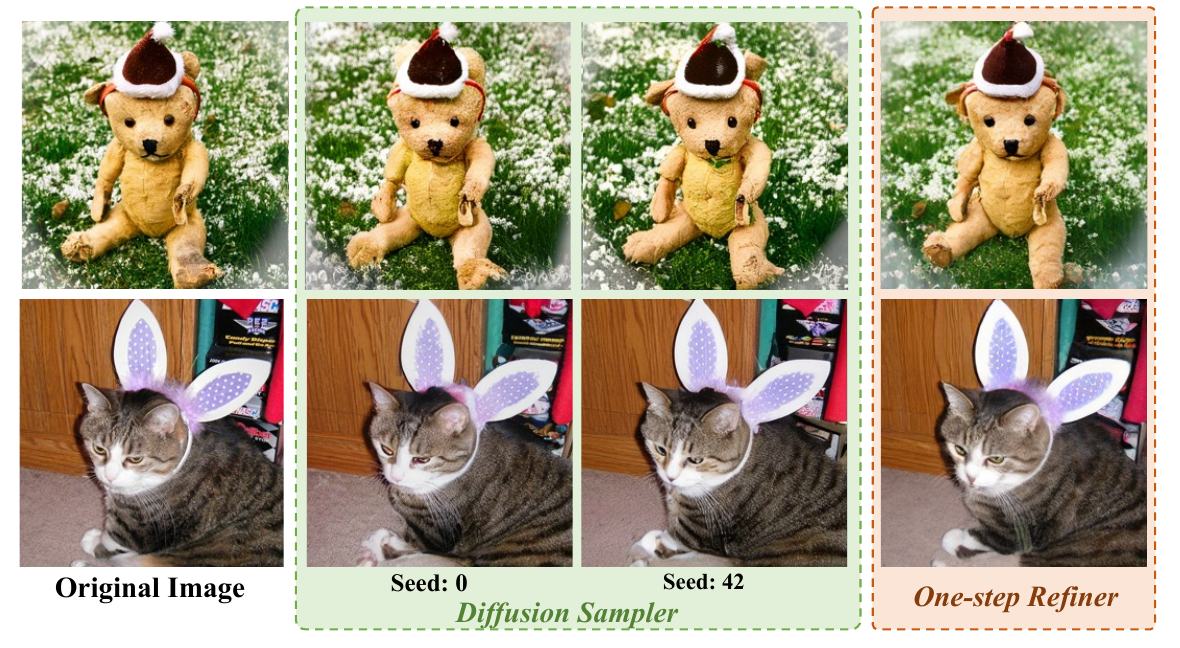}
        \caption{Reconstructions of diffusion sampler and one-step refiner.}
        \vspace{-3mm}
        \label{fig:refiner}
    \end{minipage}\\
\end{table}
% \begin{table}[t]
% \caption{Ablation on the Two-stage training.}
% \label{tab:refiner}
% \scriptsize
% \centering
% \begin{tabular}{ c |c | cccc }
% \toprule
% Tokenizer & $\#$Codebook & rFID$\downarrow$ &  PSNR $\uparrow$ & SSIM$\uparrow$ & LPIPS$\downarrow$\\
% \midrule
% Diffusion& $2^{18}$& 1.31 & 20.48 & 0.580& 0.177\\
% Refiner&$2^{18}$& \textbf{0.88} & \textbf{22.19}& \textbf{0.639} &\textbf{0.128}\\
% \midrule
% Diffusion& $2^{32}$& 1.03 & 21.43 & 0.633& 0.145\\
% Refiner& $2^{32}$&\textbf{0.67}& \textbf{23.05}& \textbf{0.684}& \textbf{0.100}\\
% \bottomrule
% \end{tabular}
% \end{table}
% \begin{figure}[t]
% \centering
% \includegraphics[width=0.9\linewidth]{figure/refiner.pdf}
% \caption{Reconstructions from diffusion sampler and one-step refiner.}   
% \label{fig:refiner}
% \end{figure}
\paragraph{Two-stage Training Strategy.}
In Tab.~\ref{tab:refiner}, we verify the effectiveness of our two-stage training strategy.
After Stage \uppercase\expandafter{\romannumeral2} training, the one-step refiner achieves consistently improved reconstruction over the initial diffusion sampler, particularly in pixel-level fidelity metrics such as PSNR and SSIM, indicating richer texture and finer details.
In Fig.~\ref{fig:refiner}, we visualize the reconstruction results of the one-step refiner and diffusion sampler.
Beyond better preserving semantics and generating finer details, the one-step refiner produces consistently realistic content while avoiding the inconsistency caused by the inherent randomness of diffusion sampling.

\section{Conclusion}
To achieve compact and semantic-rich visual representations, we proposed SemTok, a novel visual tokenizer that compresses 2D images into 1D sequences.
Through 2D-to-1D conversion, semantic alignment constraint, and generative training, SemTok delivered top-tier image reconstruction with compact compression.
Leveraging SemTok, we developed a masked AR framework for downstream image generation, with extensive experiments validating its performance in both tokenization and high-level tasks.

\section*{Impact Statement}
This paper presents work whose goal is to advance the field of Machine Learning. There are many potential societal consequences of our work, none of which we feel must be specifically highlighted here.

\bibliography{main}

@String(ICCV= {Int. Conf. Comput. Vis.})

@String(ICASSP=	{ICASSP})

@String(ICLR = {Int. Conf. Learn. Represent.})

@String(ICCV  = {ICCV})

@String(ICLR  = {ICLR})

@article{ma2025unitok,
  title={Unitok: A unified tokenizer for visual generation and understanding},
  author={Ma, Chuofan and Jiang, Yi and Wu, Junfeng and Yang, Jihan and Yu, Xin and Yuan, Zehuan and Peng, Bingyue and Qi, Xiaojuan},
  journal={arXiv preprint arXiv:2502.20321},
  year={2025}
}

@inproceedings{rombach2022high,
  title={High-resolution image synthesis with latent diffusion models},
  author={Rombach, Robin and Blattmann, Andreas and Lorenz, Dominik and Esser, Patrick and Ommer, Bj{\"o}rn},
  booktitle={Proceedings of the IEEE/CVF conference on computer vision and pattern recognition},
  pages={10684--10695},
  year={2022}
}

@inproceedings{li2023mage,
  title={Mage: Masked generative encoder to unify representation learning and image synthesis},
  author={Li, Tianhong and Chang, Huiwen and Mishra, Shlok and Zhang, Han and Katabi, Dina and Krishnan, Dilip},
  booktitle={Proceedings of the IEEE/CVF Conference on Computer Vision and Pattern Recognition},
  pages={2142--2152},
  year={2023}
}

@article{ho2020denoising,
  title={Denoising diffusion probabilistic models},
  author={Ho, Jonathan and Jain, Ajay and Abbeel, Pieter},
  journal={Advances in neural information processing systems},
  volume={33},
  pages={6840--6851},
  year={2020}
}

@article{dhariwal2021diffusion,
  title={Diffusion models beat gans on image synthesis},
  author={Dhariwal, Prafulla and Nichol, Alexander},
  journal={Advances in neural information processing systems},
  volume={34},
  pages={8780--8794},
  year={2021}
}

@article{lipman2022flow,
  title={Flow matching for generative modeling},
  author={Lipman, Yaron and Chen, Ricky TQ and Ben-Hamu, Heli and Nickel, Maximilian and Le, Matt},
  journal={arXiv preprint arXiv:2210.02747},
  year={2022}
}

@article{chen2023pixart,
  title={Pixart-$\alpha $: Fast training of diffusion transformer for photorealistic text-to-image synthesis},
  author={Chen, Junsong and Yu, Jincheng and Ge, Chongjian and Yao, Lewei and Xie, Enze and Wu, Yue and Wang, Zhongdao and Kwok, James and Luo, Ping and Lu, Huchuan and others},
  journal={arXiv preprint arXiv:2310.00426},
  year={2023}
}

@article{li2024autoregressive,
  title={Autoregressive image generation without vector quantization},
  author={Li, Tianhong and Tian, Yonglong and Li, He and Deng, Mingyang and He, Kaiming},
  journal={Advances in Neural Information Processing Systems},
  volume={37},
  pages={56424--56445},
  year={2024}
}

@article{tian2024visual,
  title={Visual autoregressive modeling: Scalable image generation via next-scale prediction},
  author={Tian, Keyu and Jiang, Yi and Yuan, Zehuan and Peng, Bingyue and Wang, Liwei},
  journal={Advances in neural information processing systems},
  volume={37},
  pages={84839--84865},
  year={2024}
}

@inproceedings{DBLP:conf/icml/ChangZBML00MFRL23,
  author       = {Huiwen Chang and
                  Han Zhang and
                  Jarred Barber and
                  Aaron Maschinot and
                  Jos{\'{e}} Lezama and
                  Lu Jiang and
                  Ming{-}Hsuan Yang and
                  Kevin Patrick Murphy and
                  William T. Freeman and
                  Michael Rubinstein and
                  Yuanzhen Li and
                  Dilip Krishnan},
  title        = {Muse: Text-To-Image Generation via Masked Generative Transformers},
  booktitle    = {{ICML}},
  series       = {Proceedings of Machine Learning Research},
  volume       = {202},
  pages        = {4055--4075},
  publisher    = {{PMLR}},
  year         = {2023}
}

@article{chen2025diffusion,
  title={Diffusion autoencoders are scalable image tokenizers},
  author={Chen, Yinbo and Girdhar, Rohit and Wang, Xiaolong and Rambhatla, Sai Saketh and Misra, Ishan},
  journal={arXiv preprint arXiv:2501.18593},
  year={2025}
}

@inproceedings{chang2022maskgit,
  title={Maskgit: Masked generative image transformer},
  author={Chang, Huiwen and Zhang, Han and Jiang, Lu and Liu, Ce and Freeman, William T},
  booktitle={Proceedings of the IEEE/CVF conference on computer vision and pattern recognition},
  pages={11315--11325},
  year={2022}
}

@article{sun2024autoregressive,
  title={Autoregressive model beats diffusion: Llama for scalable image generation},
  author={Sun, Peize and Jiang, Yi and Chen, Shoufa and Zhang, Shilong and Peng, Bingyue and Luo, Ping and Yuan, Zehuan},
  journal={arXiv preprint arXiv:2406.06525},
  year={2024}
}

@article{xie2024show,
  title={Show-o: One single transformer to unify multimodal understanding and generation},
  author={Xie, Jinheng and Mao, Weijia and Bai, Zechen and Zhang, David Junhao and Wang, Weihao and Lin, Kevin Qinghong and Gu, Yuchao and Chen, Zhijie and Yang, Zhenheng and Shou, Mike Zheng},
  journal={arXiv preprint arXiv:2408.12528},
  year={2024}
}

@article{kingma2013auto,
  title={Auto-encoding variational bayes},
  author={Kingma, Diederik P and Welling, Max},
  journal={arXiv preprint arXiv:1312.6114},
  year={2013}
}

@inproceedings{esser2021taming,
  title={Taming transformers for high-resolution image synthesis},
  author={Esser, Patrick and Rombach, Robin and Ommer, Bjorn},
  booktitle={Proceedings of the IEEE/CVF conference on computer vision and pattern recognition},
  pages={12873--12883},
  year={2021}
}

@article{yu2024image,
  title={An image is worth 32 tokens for reconstruction and generation},
  author={Yu, Qihang and Weber, Mark and Deng, Xueqing and Shen, Xiaohui and Cremers, Daniel and Chen, Liang-Chieh},
  journal={Advances in Neural Information Processing Systems},
  volume={37},
  pages={128940--128966},
  year={2024}
}

@article{wang2025selftok,
  title={Selftok: Discrete Visual Tokens of Autoregression, by Diffusion, and for Reasoning},
  author={Wang, Bohan and Yue, Zhongqi and Zhang, Fengda and Chen, Shuo and Bi, Li'an and Zhang, Junzhe and Song, Xue and Chan, Kennard Yanting and Pan, Jiachun and Wu, Weijia and others},
  journal={arXiv preprint arXiv:2505.07538},
  year={2025}
}

@article{van2017neural,
  title={Neural discrete representation learning},
  author={Van Den Oord, Aaron and Vinyals, Oriol and others},
  journal={Advances in neural information processing systems},
  volume={30},
  year={2017}
}

@article{razavi2019generating,
  title={Generating diverse high-fidelity images with vq-vae-2},
  author={Razavi, Ali and Van den Oord, Aaron and Vinyals, Oriol},
  journal={Advances in neural information processing systems},
  volume={32},
  year={2019}
}

@inproceedings{zhang2018unreasonable,
  title={The unreasonable effectiveness of deep features as a perceptual metric},
  author={Zhang, Richard and Isola, Phillip and Efros, Alexei A and Shechtman, Eli and Wang, Oliver},
  booktitle={Proceedings of the IEEE conference on computer vision and pattern recognition},
  pages={586--595},
  year={2018}
}

@article{goodfellow2020generative,
  title={Generative adversarial networks},
  author={Goodfellow, Ian and Pouget-Abadie, Jean and Mirza, Mehdi and Xu, Bing and Warde-Farley, David and Ozair, Sherjil and Courville, Aaron and Bengio, Yoshua},
  journal={Communications of the ACM},
  volume={63},
  number={11},
  pages={139--144},
  year={2020},
  publisher={ACM New York, NY, USA}
}

@article{wang2024emu3,
  title={Emu3: Next-token prediction is all you need},
  author={Wang, Xinlong and Zhang, Xiaosong and Luo, Zhengxiong and Sun, Quan and Cui, Yufeng and Wang, Jinsheng and Zhang, Fan and Wang, Yueze and Li, Zhen and Yu, Qiying and others},
  journal={arXiv preprint arXiv:2409.18869},
  year={2024}
}

@article{ge2024seed,
  title={Seed-x: Multimodal models with unified multi-granularity comprehension and generation},
  author={Ge, Yuying and Zhao, Sijie and Zhu, Jinguo and Ge, Yixiao and Yi, Kun and Song, Lin and Li, Chen and Ding, Xiaohan and Shan, Ying},
  journal={arXiv preprint arXiv:2404.14396},
  year={2024}
}

@inproceedings{bachmann2025flextok,
  title={FlexTok: Resampling Images into 1D Token Sequences of Flexible Length},
  author={Bachmann, Roman and Allardice, Jesse and Mizrahi, David and Fini, Enrico and Kar, O{\u{g}}uzhan Fatih and Amirloo, Elmira and El-Nouby, Alaaeldin and Zamir, Amir and Dehghan, Afshin},
  booktitle={Forty-second International Conference on Machine Learning},
  year={2025}
}

@inproceedings{ramesh2021zero,
  title={Zero-shot text-to-image generation},
  author={Ramesh, Aditya and Pavlov, Mikhail and Goh, Gabriel and Gray, Scott and Voss, Chelsea and Radford, Alec and Chen, Mark and Sutskever, Ilya},
  booktitle={International conference on machine learning},
  pages={8821--8831},
  year={2021},
  organization={Pmlr}
}

@article{yu2022scaling,
  title={Scaling autoregressive models for content-rich text-to-image generation},
  author={Yu, Jiahui and Xu, Yuanzhong and Koh, Jing Yu and Luong, Thang and Baid, Gunjan and Wang, Zirui and Vasudevan, Vijay and Ku, Alexander and Yang, Yinfei and Ayan, Burcu Karagol and others},
  journal={arXiv preprint arXiv:2206.10789},
  volume={2},
  number={3},
  pages={5},
  year={2022}
}

@inproceedings{lee2022autoregressive,
  title={Autoregressive image generation using residual quantization},
  author={Lee, Doyup and Kim, Chiheon and Kim, Saehoon and Cho, Minsu and Han, Wook-Shin},
  booktitle={Proceedings of the IEEE/CVF conference on computer vision and pattern recognition},
  pages={11523--11532},
  year={2022}
}

@article{zheng2022movq,
  title={Movq: Modulating quantized vectors for high-fidelity image generation},
  author={Zheng, Chuanxia and Vuong, Tung-Long and Cai, Jianfei and Phung, Dinh},
  journal={Advances in Neural Information Processing Systems},
  volume={35},
  pages={23412--23425},
  year={2022}
}

@article{mentzer2023finite,
  title={Finite scalar quantization: Vq-vae made simple},
  author={Mentzer, Fabian and Minnen, David and Agustsson, Eirikur and Tschannen, Michael},
  journal={arXiv preprint arXiv:2309.15505},
  year={2023}
}

@article{kynkaanniemi2019improved,
  title={Improved precision and recall metric for assessing generative models},
  author={Kynk{\"a}{\"a}nniemi, Tuomas and Karras, Tero and Laine, Samuli and Lehtinen, Jaakko and Aila, Timo},
  journal={Advances in neural information processing systems},
  volume={32},
  year={2019}
}

@article{zhao2024image,
  title={Image and video tokenization with binary spherical quantization},
  author={Zhao, Yue and Xiong, Yuanjun and Kr{\"a}henb{\"u}hl, Philipp},
  journal={arXiv preprint arXiv:2406.07548},
  year={2024}
}

@article{luo2024open,
  title={Open-magvit2: An open-source project toward democratizing auto-regressive visual generation},
  author={Luo, Zhuoyan and Shi, Fengyuan and Ge, Yixiao and Yang, Yujiu and Wang, Limin and Shan, Ying},
  journal={arXiv preprint arXiv:2409.04410},
  year={2024}
}

@article{agarwal2025cosmos,
  title={Cosmos world foundation model platform for physical ai},
  author={Agarwal, Niket and Ali, Arslan and Bala, Maciej and Balaji, Yogesh and Barker, Erik and Cai, Tiffany and Chattopadhyay, Prithvijit and Chen, Yongxin and Cui, Yin and Ding, Yifan and others},
  journal={arXiv preprint arXiv:2501.03575},
  year={2025}
}

@article{sargent2025flow,
  title={Flow to the mode: Mode-seeking diffusion autoencoders for state-of-the-art image tokenization},
  author={Sargent, Kyle and Hsu, Kyle and Johnson, Justin and Fei-Fei, Li and Wu, Jiajun},
  journal={arXiv preprint arXiv:2503.11056},
  year={2025}
}

@article{yu2021vector,
  title={Vector-quantized image modeling with improved vqgan},
  author={Yu, Jiahui and Li, Xin and Koh, Jing Yu and Zhang, Han and Pang, Ruoming and Qin, James and Ku, Alexander and Xu, Yuanzhong and Baldridge, Jason and Wu, Yonghui},
  journal={arXiv preprint arXiv:2110.04627},
  year={2021}
}

@article{yu2023language,
  title={Language Model Beats Diffusion--Tokenizer is Key to Visual Generation},
  author={Yu, Lijun and Lezama, Jos{\'e} and Gundavarapu, Nitesh B and Versari, Luca and Sohn, Kihyuk and Minnen, David and Cheng, Yong and Birodkar, Vighnesh and Gupta, Agrim and Gu, Xiuye and others},
  journal={arXiv preprint arXiv:2310.05737},
  year={2023}
}

@inproceedings{radford2021learning,
  title={Learning transferable visual models from natural language supervision},
  author={Radford, Alec and Kim, Jong Wook and Hallacy, Chris and Ramesh, Aditya and Goh, Gabriel and Agarwal, Sandhini and Sastry, Girish and Askell, Amanda and Mishkin, Pamela and Clark, Jack and others},
  booktitle={International conference on machine learning},
  pages={8748--8763},
  year={2021},
  organization={PmLR}
}

@inproceedings{li2023blip,
  title={Blip-2: Bootstrapping language-image pre-training with frozen image encoders and large language models},
  author={Li, Junnan and Li, Dongxu and Savarese, Silvio and Hoi, Steven},
  booktitle={International conference on machine learning},
  pages={19730--19742},
  year={2023},
  organization={PMLR}
}

@inproceedings{jansen2020coincidence,
  title={Coincidence, categorization, and consolidation: Learning to recognize sounds with minimal supervision},
  author={Jansen, Aren and Ellis, Daniel PW and Hershey, Shawn and Moore, R Channing and Plakal, Manoj and Popat, Ashok C and Saurous, Rif A},
  booktitle={ICASSP 2020-2020 IEEE International Conference on Acoustics, Speech and Signal Processing (ICASSP)},
  pages={121--125},
  year={2020},
  organization={IEEE}
}

@article{su2024roformer,
  title={Roformer: Enhanced transformer with rotary position embedding},
  author={Su, Jianlin and Ahmed, Murtadha and Lu, Yu and Pan, Shengfeng and Bo, Wen and Liu, Yunfeng},
  journal={Neurocomputing},
  volume={568},
  pages={127063},
  year={2024},
  publisher={Elsevier}
}

@inproceedings{karras2019style,
  title={A style-based generator architecture for generative adversarial networks},
  author={Karras, Tero and Laine, Samuli and Aila, Timo},
  booktitle={Proceedings of the IEEE/CVF conference on computer vision and pattern recognition},
  pages={4401--4410},
  year={2019}
}

@inproceedings{DBLP:conf/iclr/LiuG023,
  author       = {Xingchao Liu and
                  Chengyue Gong and
                  Qiang Liu},
  title        = {Flow Straight and Fast: Learning to Generate and Transfer Data with
                  Rectified Flow},
  booktitle    = {{ICLR}},
  publisher    = {OpenReview.net},
  year         = {2023}
}

@article{liu2023visual,
  title={Visual instruction tuning},
  author={Liu, Haotian and Li, Chunyuan and Wu, Qingyang and Lee, Yong Jae},
  journal={Advances in neural information processing systems},
  volume={36},
  pages={34892--34916},
  year={2023}
}

@article{sun2023emu,
  title={Emu: Generative pretraining in multimodality},
  author={Sun, Quan and Yu, Qiying and Cui, Yufeng and Zhang, Fan and Zhang, Xiaosong and Wang, Yueze and Gao, Hongcheng and Liu, Jingjing and Huang, Tiejun and Wang, Xinlong},
  journal={arXiv preprint arXiv:2307.05222},
  year={2023}
}

@inproceedings{DBLP:conf/iclr/Jin0XCLTHCSMZOG24,
  author       = {Yang Jin and
                  Kun Xu and
                  Kun Xu and
                  Liwei Chen and
                  Chao Liao and
                  Jianchao Tan and
                  Quzhe Huang and
                  Bin Chen and
                  Chengru Song and
                  Dai Meng and
                  Di Zhang and
                  Wenwu Ou and
                  Kun Gai and
                  Yadong Mu},
  title        = {Unified Language-Vision Pretraining in {LLM} with Dynamic Discrete
                  Visual Tokenization},
  booktitle    = {{ICLR}},
  publisher    = {OpenReview.net},
  year         = {2024}
}

@article{zheng2025diffusion,
  title={Diffusion Transformers with Representation Autoencoders},
  author={Zheng, Boyang and Ma, Nanye and Tong, Shengbang and Xie, Saining},
  journal={arXiv preprint arXiv:2510.11690},
  year={2025}
}

@inproceedings{yao2025reconstruction,
  title={Reconstruction vs. generation: Taming optimization dilemma in latent diffusion models},
  author={Yao, Jingfeng and Yang, Bin and Wang, Xinggang},
  booktitle={Proceedings of the Computer Vision and Pattern Recognition Conference},
  pages={15703--15712},
  year={2025}
}

@article{zhao2025qlip,
  title={Qlip: Text-aligned visual tokenization unifies auto-regressive multimodal understanding and generation},
  author={Zhao, Yue and Xue, Fuzhao and Reed, Scott and Fan, Linxi and Zhu, Yuke and Kautz, Jan and Yu, Zhiding and Kr{\"a}henb{\"u}hl, Philipp and Huang, De-An},
  journal={arXiv preprint arXiv:2502.05178},
  year={2025}
}

@inproceedings{DBLP:conf/nips/DhariwalN21,
  author       = {Prafulla Dhariwal and
                  Alexander Quinn Nichol},
  title        = {Diffusion Models Beat GANs on Image Synthesis},
  booktitle    = {NeurIPS},
  pages        = {8780--8794},
  year         = {2021}
}

@inproceedings{DBLP:conf/icml/HoogeboomHS23,
  author       = {Emiel Hoogeboom and
                  Jonathan Heek and
                  Tim Salimans},
  title        = {simple diffusion: End-to-end diffusion for high resolution images},
  booktitle    = {{ICML}},
  series       = {Proceedings of Machine Learning Research},
  volume       = {202},
  pages        = {13213--13232},
  publisher    = {{PMLR}},
  year         = {2023}
}

@inproceedings{kang2023scaling,
  title={Scaling up gans for text-to-image synthesis},
  author={Kang, Minguk and Zhu, Jun-Yan and Zhang, Richard and Park, Jaesik and Shechtman, Eli and Paris, Sylvain and Park, Taesung},
  booktitle={Proceedings of the IEEE/CVF conference on computer vision and pattern recognition},
  pages={10124--10134},
  year={2023}
}

@article{brock2018large,
  title={Large scale GAN training for high fidelity natural image synthesis},
  author={Brock, Andrew and Donahue, Jeff and Simonyan, Karen},
  journal={arXiv preprint arXiv:1809.11096},
  year={2018}
}

@inproceedings{DBLP:conf/nips/KingmaG23,
  author       = {Diederik P. Kingma and
                  Ruiqi Gao},
  title        = {Understanding Diffusion Objectives as the {ELBO} with Simple Data
                  Augmentation},
  booktitle    = {NeurIPS},
  year         = {2023}
}

@article{salimans2016improved,
  title={Improved techniques for training gans},
  author={Salimans, Tim and Goodfellow, Ian and Zaremba, Wojciech and Cheung, Vicki and Radford, Alec and Chen, Xi},
  journal={Advances in neural information processing systems},
  volume={29},
  year={2016}
}

@inproceedings{he2022masked,
  title={Masked autoencoders are scalable vision learners},
  author={He, Kaiming and Chen, Xinlei and Xie, Saining and Li, Yanghao and Doll{\'a}r, Piotr and Girshick, Ross},
  booktitle={Proceedings of the IEEE/CVF conference on computer vision and pattern recognition},
  pages={16000--16009},
  year={2022}
}

@article{ho2022classifier,
  title={Classifier-free diffusion guidance},
  author={Ho, Jonathan and Salimans, Tim},
  journal={arXiv preprint arXiv:2207.12598},
  year={2022}
}

@article{heusel2017gans,
  title={Gans trained by a two time-scale update rule converge to a local nash equilibrium},
  author={Heusel, Martin and Ramsauer, Hubert and Unterthiner, Thomas and Nessler, Bernhard and Hochreiter, Sepp},
  journal={Advances in neural information processing systems},
  volume={30},
  year={2017}
}

@inproceedings{deng2009imagenet,
  title={Imagenet: A large-scale hierarchical image database},
  author={Deng, Jia and Dong, Wei and Socher, Richard and Li, Li-Jia and Li, Kai and Fei-Fei, Li},
  booktitle={2009 IEEE conference on computer vision and pattern recognition},
  pages={248--255},
  year={2009},
  organization={Ieee}
}

@inproceedings{DBLP:conf/iccv/PeeblesX23,
  author       = {William Peebles and
                  Saining Xie},
  title        = {Scalable Diffusion Models with Transformers},
  booktitle    = {{ICCV}},
  pages        = {4172--4182},
  publisher    = {{IEEE}},
  year         = {2023}
}

@inproceedings{DBLP:conf/iclr/YangTZ00XYHZFYZ25,
  author       = {Zhuoyi Yang and
                  Jiayan Teng and
                  Wendi Zheng and
                  Ming Ding and
                  Shiyu Huang and
                  Jiazheng Xu and
                  Yuanming Yang and
                  Wenyi Hong and
                  Xiaohan Zhang and
                  Guanyu Feng and
                  Da Yin and
                  Yuxuan Zhang and
                  Weihan Wang and
                  Yean Cheng and
                  Bin Xu and
                  Xiaotao Gu and
                  Yuxiao Dong and
                  Jie Tang},
  title        = {CogVideoX: Text-to-Video Diffusion Models with An Expert Transformer},
  booktitle    = {{ICLR}},
  publisher    = {OpenReview.net},
  year         = {2025}
}

@inproceedings{han2025infinity,
  title={Infinity: Scaling bitwise autoregressive modeling for high-resolution image synthesis},
  author={Han, Jian and Liu, Jinlai and Jiang, Yi and Yan, Bin and Zhang, Yuqi and Yuan, Zehuan and Peng, Bingyue and Liu, Xiaobing},
  booktitle={Proceedings of the Computer Vision and Pattern Recognition Conference},
  pages={15733--15744},
  year={2025}
}

@inproceedings{pang2025randar,
  title={Randar: Decoder-only autoregressive visual generation in random orders},
  author={Pang, Ziqi and Zhang, Tianyuan and Luan, Fujun and Man, Yunze and Tan, Hao and Zhang, Kai and Freeman, William T and Wang, Yu-Xiong},
  booktitle={Proceedings of the Computer Vision and Pattern Recognition Conference},
  pages={45--55},
  year={2025}
}

@InProceedings{Yu_2025_ICCV,
    author    = {Yu, Qihang and He, Ju and Deng, Xueqing and Shen, Xiaohui and Chen, Liang-Chieh},
    title     = {Randomized Autoregressive Visual Generation},
    booktitle = {Proceedings of the IEEE/CVF International Conference on Computer Vision (ICCV)},
    month     = {October},
    year      = {2025},
    pages     = {18431-18441}
}

@inproceedings{DBLP:conf/iclr/YuLKZPQKXBW22,
  author       = {Jiahui Yu and
                  Xin Li and
                  Jing Yu Koh and
                  Han Zhang and
                  Ruoming Pang and
                  James Qin and
                  Alexander Ku and
                  Yuanzhong Xu and
                  Jason Baldridge and
                  Yonghui Wu},
  title        = {Vector-quantized Image Modeling with Improved {VQGAN}},
  booktitle    = {{ICLR}},
  publisher    = {OpenReview.net},
  year         = {2022}
}

@inproceedings{he2016deep,
  title={Deep residual learning for image recognition},
  author={He, Kaiming and Zhang, Xiangyu and Ren, Shaoqing and Sun, Jian},
  booktitle={Proceedings of the IEEE conference on computer vision and pattern recognition},
  pages={770--778},
  year={2016}
}

@article{DBLP:journals/tmlr/WeberYYDSCC24,
  author       = {Mark Weber and
                  Lijun Yu and
                  Qihang Yu and
                  Xueqing Deng and
                  Xiaohui Shen and
                  Daniel Cremers and
                  Liang{-}Chieh Chen},
  title        = {MaskBit: Embedding-free Image Generation via Bit Tokens},
  journal      = {Trans. Mach. Learn. Res.},
  volume       = {2024},
  year         = {2024}
}

@inproceedings{DBLP:conf/naacl/DevlinCLT19,
  author       = {Jacob Devlin and
                  Ming{-}Wei Chang and
                  Kenton Lee and
                  Kristina Toutanova},
  title        = {{BERT:} Pre-training of Deep Bidirectional Transformers for Language
                  Understanding},
  booktitle    = {{NAACL-HLT} {(1)}},
  pages        = {4171--4186},
  publisher    = {Association for Computational Linguistics},
  year         = {2019}
}

@inproceedings{esser2024scaling,
  title={Scaling rectified flow transformers for high-resolution image synthesis},
  author={Esser, Patrick and Kulal, Sumith and Blattmann, Andreas and Entezari, Rahim and M{\"u}ller, Jonas and Saini, Harry and Levi, Yam and Lorenz, Dominik and Sauer, Axel and Boesel, Frederic and others},
  booktitle={Forty-first international conference on machine learning},
  year={2024}
}

@article{tschannen2025siglip,
  title={Siglip 2: Multilingual vision-language encoders with improved semantic understanding, localization, and dense features},
  author={Tschannen, Michael and Gritsenko, Alexey and Wang, Xiao and Naeem, Muhammad Ferjad and Alabdulmohsin, Ibrahim and Parthasarathy, Nikhil and Evans, Talfan and Beyer, Lucas and Xia, Ye and Mustafa, Basil and others},
  journal={arXiv preprint arXiv:2502.14786},
  year={2025}
}

@inproceedings{dehghani2023scaling,
  title={Scaling vision transformers to 22 billion parameters},
  author={Dehghani, Mostafa and Djolonga, Josip and Mustafa, Basil and Padlewski, Piotr and Heek, Jonathan and Gilmer, Justin and Steiner, Andreas Peter and Caron, Mathilde and Geirhos, Robert and Alabdulmohsin, Ibrahim and others},
  booktitle={International conference on machine learning},
  pages={7480--7512},
  year={2023},
  organization={PMLR}
}

@inproceedings{zhai2023sigmoid,
  title={Sigmoid loss for language image pre-training},
  author={Zhai, Xiaohua and Mustafa, Basil and Kolesnikov, Alexander and Beyer, Lucas},
  booktitle={Proceedings of the IEEE/CVF international conference on computer vision},
  pages={11975--11986},
  year={2023}
}

@article{bengio2013estimating,
  title={Estimating or propagating gradients through stochastic neurons for conditional computation},
  author={Bengio, Yoshua and L{\'e}onard, Nicholas and Courville, Aaron},
  journal={arXiv preprint arXiv:1308.3432},
  year={2013}
}
\bibliographystyle{icml2026}

%%%%%%%%%%%%%%%%%%%%%%%%%%%%%%%%%%%%%%%%%%%%%%%%%%%%%%%%%%%%%%%%%%%%%%%%%%%%%%%
%%%%%%%%%%%%%%%%%%%%%%%%%%%%%%%%%%%%%%%%%%%%%%%%%%%%%%%%%%%%%%%%%%%%%%%%%%%%%%%
% APPENDIX
%%%%%%%%%%%%%%%%%%%%%%%%%%%%%%%%%%%%%%%%%%%%%%%%%%%%%%%%%%%%%%%%%%%%%%%%%%%%%%%
%%%%%%%%%%%%%%%%%%%%%%%%%%%%%%%%%%%%%%%%%%%%%%%%%%%%%%%%%%%%%%%%%%%%%%%%%%%%%%%
\newpage
\appendix
\onecolumn
\section{Implementation Details}

\subsection{Details of the SemTok Tokenizer}
\begin{table}[h]
\caption{Architecture Configuration of SemTok.}
\label{tab:semtok_arch}
\scriptsize
\centering
\begin{tabular}{ c | cccc }
\toprule
Model & Depth & Width  & Heads & \# Params\\
\midrule
Encoder & 8 & 768 & 12 & 113M\\
Decoder & 24 & 1536 & 24 & 2236M\\
\bottomrule
\end{tabular}
\end{table}
\paragraph{Encoder.} Our encoder follows an MMDiT-based \cite{esser2024scaling} architecture. 
Drawing on the findings in FlowMo \cite{sargent2025flow}, we argue that a lightweight encoder is sufficient to accomplish the compression and representation of image semantics, where the configuration is listed in \ref{tab:semtok_arch}.
Images are first compressed into latents using SD3.5-VAE \cite{rombach2022high} with a downscale factor of 8.
Both the encoder and decoder operate on $2\times 2$ patches of VAE latents to reduce the sequence lengths.
%The compressed latent patches are fed into the image branch of the dual-branch architecture, while the token branch is supplied with learnable embeddings of length 256 as the mask sequence to be encoded.
We apply 2D RoPE \cite{su2024roformer} and 1D RoPE to represent positional relationships of 2D image tokens and 1D semantic tokens, respectively.

\paragraph{Decoder.} We argue that, as a generator, the decoder is tasked with reconstructing raw pixels from the compressed sequence, and thus requires a large architecture to learn the generation prior.
We thus adopt SD3.5 \cite{esser2024scaling} for the decoder and initialize it with its pre-trained parameters to inherit robust generation capabilities, where the configuration is listed in \ref{tab:semtok_arch}.
Consistent with the encoder, we apply 2D and 1D RoPE to the two branches of the decoder as well. 
%During the diffusion-based pre-training phase, noisy latents are fed into the 2D image branch for $v$-prediction, whereas in the refinement-orient fine-tuning phase, learnable mask sequences are input to the 2D branch.
The predicted latents are mapped back to raw pixels via the VAE decoder.

\paragraph{Other Configuration.}
For data augmentation, we resize the input image to $256\times256$ and apply horizontal flipping.
In Stage \uppercase\expandafter{\romannumeral1}, we utilize the uniform noise distribution schedule on $[0,1]$ for rectified flow training.
During inference, SemTok predicts images from noise through 25-step iteration with a sampler in \cite{esser2024scaling}.
In Stage \uppercase\expandafter{\romannumeral2}, we adopt the same discriminator as that in \cite{DBLP:journals/tmlr/WeberYYDSCC24}, which is incorporated after 20,000 iterations with the learning rate set to $1\times10^{-4}$.
The learning rate is warmed up over the first 10,000 iterations and scheduled following cosine annealing.
We use the exponential moving average (EMA) with a rate of 0.9999.
In all ablation experiments, we employ SemTok with a codebook size of $2^{18}$ to accelerate the convergence rate under limited computational resources.

\subsection{Details of the Autoregressive Framework}
\begin{table}[h]
\caption{Architecture Configuration of the AR model.}
\label{tab:ar_arch}
\scriptsize
\centering
\begin{tabular}{ c | cccc }
\toprule
Size & Depth & Width  & Heads & \# Params\\
\midrule
L & 24 & 1024 & 16 & 318M\\
XL & 36 & 1280 & 20 & 746M\\
XXL & 24 & 2048 & 32 & 1.2B\\
\bottomrule
\end{tabular}
\end{table}

\paragraph{Model Architecture.}
We employ SemTok with a codebook size of $2^{18}$ as the basic visual tokenizer.
Our AR model follows the architecture of LlamaGen \cite{sun2024autoregressive} with 1D RoPE \cite{su2024roformer}.
Our model employs a bidirectional attention mechanism for global correlations, unlike the original model with a causal mask.
We employ QK normalization \cite{dehghani2023scaling} to stabilize training.
In Tab.~\ref{tab:ar_arch}, we scale the model size from 318M up to 1.2B parameters following previous works \cite{sun2024autoregressive}. 
For class-conditioned image generation, we concatenate the learnable class token with the image tokens and place it at the beginning. 
Additionally, it serves as a condition through AdaLN \cite{DBLP:conf/iccv/PeeblesX23}.
Given the large codebook size of our tokenizer to enhance its representational capacity, a conventional classifier would demand substantial storage space.
We therefore adopt the Infinite-Vocabulary Classifier (IVC) in \cite{han2025infinity}, which performs binary classification on each index bit to accelerate convergence.

\paragraph{Inference.}
Our AR model predicts several tokens from masked tokens progressively in a fully randomized order, with the masking ratio following a cosine schedule.
This enables us to predict multiple tokens in parallel at each step, with 64 steps set as the default to generate 256 tokens.
We adopt top-$p$ based filtering techniques, since our bitwise classifier only supports binary sampling for each bit.
We sweep the optimal top-$p$ value and set it to 0.75.

\paragraph{Other Configuration.}
We apply a dropout probability of 0.1 on the class condition to adopt CFG \cite{ho2022classifier}. 
We use a simple linear CFG schedule, where the scale rises linearly to its maximum based on the ratio of predicted tokens. 
We sweep the optimal guidance scale for each model with 6.0 for SemTok-AR-L, 5.3 for SemTok-AR-XL, and 4.9 for SemTok-AR-XXL.
We adopt label smoothing with a factor of 0.01 to enhance the generalization.
%The AR model also adopts EMA and learning rate scheduling strategies similar to those of SemTok.
In all ablation experiments, we employ AR model-L to accelerate the convergence rate under limited computational resources.

\section{Additional Experimental Results}

\subsection{Ablation Study of SemTok Tokenizer}
\begin{table}[h]
\caption{Ablation on the \textit{Quantizer}.}
\label{tab:quant}
\scriptsize
\centering
\begin{tabular}{ c | cccc }
\toprule
Quantizer & rFID$\downarrow$ &  PSNR $\uparrow$ & SSIM$\uparrow$ & LPIPS$\downarrow$\\
\midrule
LFQ \cite{yu2023language}& 1.32 & 20.78 & 0.564 & 0.163\\
BSQ \cite{zhao2024image} & \textbf{0.88} & \textbf{22.19}& \textbf{0.639} &\textbf{0.128}\\
\bottomrule
\end{tabular}
\end{table}
\paragraph{\textbf{\textit{Binary Spherical Quantizer.}}}
Our SemTok adopts BSQ \cite{zhao2024image} as the quantizer, which projects the high-dimensional visual embedding to a lower-dimensional hypersphere and applies binary quantization without requiring extra space for codebook storage in VQ \cite{van2017neural}.
Compared with LFQ \cite{yu2023language}, BSQ enables efficient entropy calculation via soft quantization and bounds quantization errors through $\mathcal{L}_2$ normalization, thus achieving faster and better convergence.
As shown in Tab.~\ref{tab:quant}, BSQ achieves superior results in image reconstruction tasks, verifying its stronger generalization and adaptability.

\begin{table}[h]
\caption{Ablation on the \textit{Perceptual Loss}.}
\label{tab:perc}
\scriptsize
\centering
\begin{tabular}{ c | cccc }
\toprule
$\mathcal{L}_{\textit{per}}$ & rFID$\downarrow$ &  PSNR $\uparrow$ & SSIM$\uparrow$ & LPIPS$\downarrow$\\
\midrule
LPIPS \cite{zhang2018unreasonable}& 1.13 & \textbf{22.73} & \textbf{0.649} & 0.133 \\
ResNet \cite{he2016deep} & 0.89 & 22.02 & 0.622 & 0.133 \\
\midrule
LPIPS + ResNet & \textbf{0.88} & 22.19& 0.639 &\textbf{0.128}\\
\bottomrule
\end{tabular}
\end{table}
\paragraph{\textbf{\textit{Perceptual Loss.}}}
Previous works \cite{DBLP:journals/tmlr/WeberYYDSCC24, sargent2025flow} verify the effectiveness of ResNet-based \cite{he2016deep} perceptual loss for image tokenization.
Compared with LPIPS \cite{zhang2018unreasonable}, which uses intermediate feature distance loss to enhance the similarity between local features, the ResNet loss based on logits is more capable of focusing on the fidelity of high-level semantics in reconstruction.
We consider both perspectives important for visual tokenization. Thus, unlike \cite{DBLP:journals/tmlr/WeberYYDSCC24, sargent2025flow}, we sum these two losses as the overall perceptual loss.
In Tab.~\ref{tab:perc}, the joint application of both losses yields superior results compared with using a single loss.
Furthermore, LPIPS loss focuses on improving local fidelity metrics such as SSIM, while ResNet loss targets metrics like rFID that reflect semantic distribution similarity, which aligns with our expectations.

\begin{table}[h]
\caption{Ablation on the \textit{Generative Training Strategy}.}
\label{tab:training}
\scriptsize
\centering
\begin{tabular}{ c | cccc }
\toprule
Stage & rFID$\downarrow$ &  PSNR $\uparrow$ & SSIM$\uparrow$ & LPIPS$\downarrow$\\
\midrule
Only Fine-tuning & 1.80 & 20.54 & 0.556 & 0.177\\
Pre-training \& Fine-tuning & \textbf{0.88} & \textbf{22.19}& \textbf{0.639} &\textbf{0.128}\\
\bottomrule
\end{tabular}
\end{table}
\paragraph{\textbf{\textit{Generative Training Strategy.}}}
Our SemTok utilizes a two-stage generative training strategy to explore a richer semantic distribution in the latent space.
Specifically, Stage \uppercase\expandafter{\romannumeral1} drives generative space exploration via a diffusion-based decoder, which searches for optimal sampling paths from randomly distributed noise to achieve superior semantic representation. 
In contrast, Stage \uppercase\expandafter{\romannumeral2} fine-tuning aligns with the classical pixel-level reconstruction pipeline, focusing on detail refinement rather than further modeling of semantic distribution.
In Tab.~\ref{tab:training}, models incorporating generative pre-training yield significantly superior performance compared to only fine-tuning.
Tab.~\ref{tab:refiner} and Tab.~\ref{tab:training} show that Stage \uppercase\expandafter{\romannumeral1} enhances semantic understanding, while Stage \uppercase\expandafter{\romannumeral2} ensures local detail fidelity. 
Their synergy resolves the semantic-detail trade-off, ultimately leading to enhanced performance in visual tokenization tasks.

% Quantitative metrics in Tab.~\ref{tab:training} and Tab.~\ref{refiner} demonstrate that the generative pre-training in Stage \uppercase\expandafter{\romannumeral1} equips the model with a more comprehensive understanding of semantic distributions, while the pixel-level fine-tuning in Stage \uppercase\expandafter{\romannumeral2} ensures the fidelity of local details. 
% The synergy between these two stages addresses the trade-off between semantic richness and detail precision that plagues single-stage training methods, ultimately leading to enhanced performance in downstream visual tokenization tasks.

\subsection{Ablation Study of Autoregressive Framework}

\begin{figure}[h]
\centering
\includegraphics[width=0.66\linewidth]{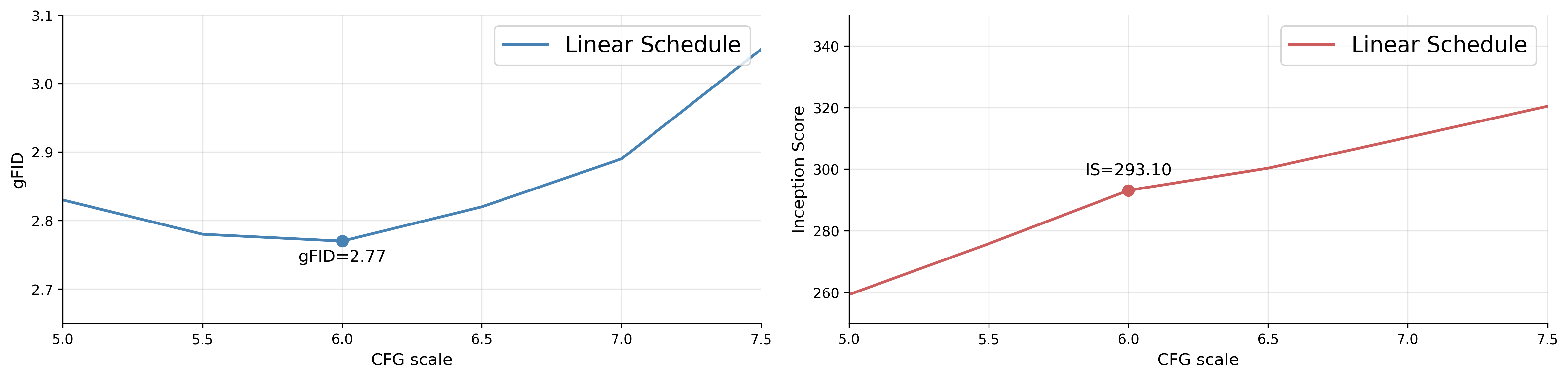}
\caption{The influence of CFG scale under \textit{Linear Schedule}.}   
\label{fig:cfg_linear}
\end{figure}
\paragraph{\textbf{\textit{CFG Scale.}}}
Our AR model incorporates CFG \cite{ho2022classifier} to boost generation quality, which imposes guidance on the sampling procedure, thereby shifting the distribution toward more desirable characteristics.
We dynamically adjust the CFG scale via a linear schedule during iterative sampling, and identify its optimal value through systematic search over predefined candidates, which is shown in Fig.~\ref{fig:cfg_linear}.
Increasing the CFG scale continuously improves generation quality (\textit{i.e.}, Inception Score), whereas its impact on gFID shows a non-monotonic trend of initial decline then enhancement.

\begin{figure}[h]
\centering
\includegraphics[width=0.66\linewidth]{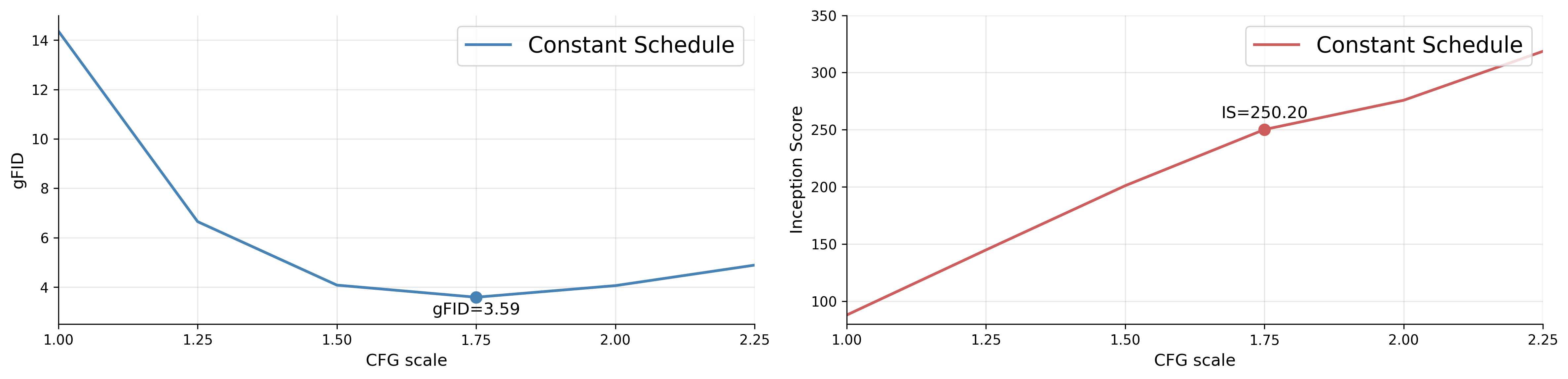}
\caption{The influence of CFG scale under \textit{Constant Schedule}.}   
\label{fig:cfg_constant}
\end{figure}
\paragraph{\textbf{\textit{CFG Schedule.}}}
We compare the linear schedule of CFG against the classic constant schedule, with the search results for the latter presented in Fig.~\ref{fig:cfg_constant} (scale=1.0 denotes without CFG).
The variation trend of the constant schedule is similar to that of the linear schedule, but its optimal performance is inferior.
We argue that a gradual increase schedule in CFG scale during iterations is more optimal, since the autoregressive process suffers from insufficient prior information in its early stage, where an excessively large CFG scale may cause distributional over-shifting or collapse.

\begin{table}[h]
    \scriptsize
  \centering
  \caption{Ablation on the \textit{Bitwise Classifier}.}
  \label{tab:bit}
  \begin{tabular}{c|cccc}
    \toprule
    Groups of Codebook & gFID$\downarrow$ &  IS $\uparrow$ & Pre$\uparrow$ & Rec$\uparrow$ \\ 
    \midrule
2 & 4.77 & 251.4 & 0.76 & 0.48\\
18 (Bitwise) & \textbf{2.77}& \textbf{293.1} & \textbf{0.78}& \textbf{0.60}\\
\bottomrule
  \end{tabular}
\end{table} 
\paragraph{\textbf{\textit{Bitwise Classifier.}}}
We use a bitwise classifier to predict bit labels, equivalent to partitioning the codebook with size $|\mathcal{C}|$ into $\log_2{|\mathcal{C}|}$ groups. 
We compare this approach with the classifier that partitions the codebook into only 2 groups and predicts index labels under a codebook size of $2^{18}$, since a large vocabulary size would lead to excessive memory overhead for conventional classifiers.
As shown in Tab.~\ref{tab:bit}, the bit-wise classifier achieves superior generation performance and faster convergence.
We argue that this effectiveness to the BSQ of the codebook. 
As analyzed in \cite{DBLP:journals/tmlr/WeberYYDSCC24}, BSQ imparts structured semantics to bit-token representations: adjacent tokens (Hamming distance = 1) are semantically similar, and flipping bits does not induce substantial semantic changes.

\begin{table}[h]
    \scriptsize
  \centering
  \caption{Ablation on the \textit{Sampling Strategy}.}
  \label{tab:sampling}
  \begin{tabular}{c|cccc}
    \toprule
    Strategy & gFID$\downarrow$ &  IS $\uparrow$ & Pre$\uparrow$ & Rec$\uparrow$ \\ 
    \midrule
Confidence-based & 3.60 & 246.8& 0.74&\textbf{0.62}\\
Random & \textbf{2.77}& \textbf{293.1} & \textbf{0.78}& 0.60\\
\bottomrule
  \end{tabular}
\end{table} 
\paragraph{\textbf{\textit{Sampling Strategy.}}}
Our AR model follows the same design as MAR \cite{li2024autoregressive}, employing a fully randomized order to predict tokens at different positions.
This randomized order differs from that of masked autoregressive methods such as MaskGIT \cite{chang2022maskgit} and MAGE \cite{li2023mage}, where the positions of the next tokens to predict are dynamically determined by their confidence.
In other words, this confidence-based sampling strategy prioritizes predicting tokens with higher confidence.
As shown in Tab.~\ref{tab:sampling}, the random sampling strategy achieves better gFID performance, as its random order induces more diverse prediction distributions.

\begin{figure}[h]
\centering
\includegraphics[width=0.66\linewidth]{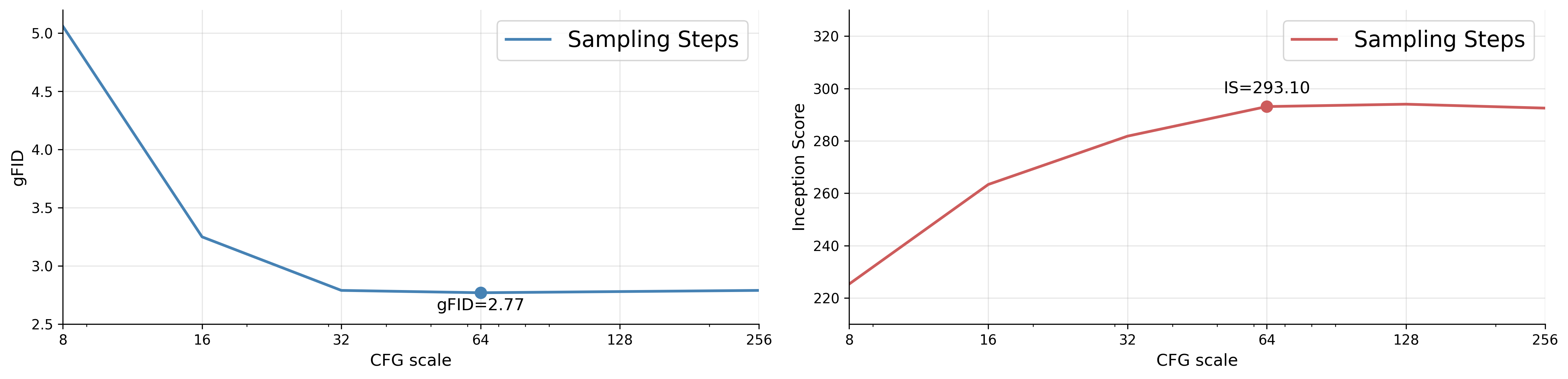} 
\caption{Ablation on the \textit{Sampling Steps} for token generation.}   
\label{fig:gen_step}
\end{figure}
\paragraph{\textbf{\textit{Sampling Steps.}}}
AR models based on next-token prediction can only generate one token per inference step, making the number of steps a critical factor affecting computational efficiency.
Our masked autoregressive modeling enables bidirectional token dependencies, allowing multiple tokens to be generated in parallel.
Thus, we can improve generation speed by reducing sampling steps.
As shown in Fig.~\ref{fig:gen_step}, once the sampling steps reach 64, additional steps yield only marginal improvements.
We argue that 64 steps represent an optimal trade-off between speed and accuracy.
%As shown in Fig.~\ref{fig:gen_step}, the reduction in sampling steps has minimal impact on generation quality, and we argue that 64 steps represent an optimal trade-off between speed and accuracy.

\begin{figure}[h]
\centering
\includegraphics[width=0.66\linewidth]{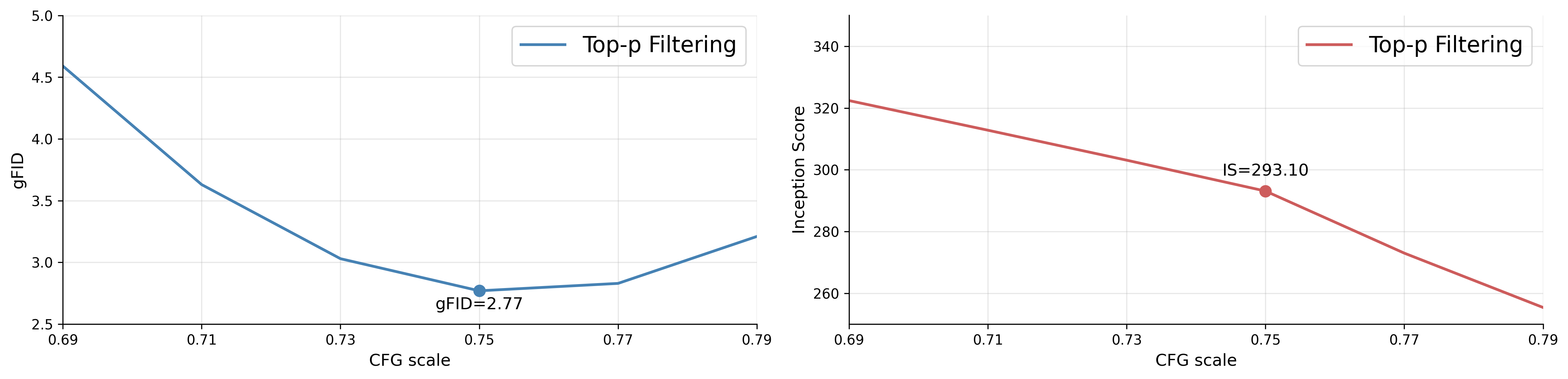} 
\caption{Ablation on the \textit{Top-$p$ Filtering} for token generation.}   
\label{fig:topp}
\end{figure}
\paragraph{\textbf{\textit{Top-$\textbf{p}$ Filtering.}}}
Based on our bitwise classifier, we apply a top‑$p$ based filtering strategy to each bit of the vector index, retaining only those bit categories whose cumulative probability exceeds $p$.
A larger $p$ ensures generation diversity, while a smaller $p$ guarantees image quality.
As shown in Fig.~\ref{fig:topp}, we find that $p=0.75$ strikes the best balance with the lowest gFid.
We use the same hyperparameter settings across all experiments.

\subsection{Limitations}
In this work, we propose SemTok, a novel semantic one-dimensional tokenizer that compresses 2D imlimitation 1D discrete tokens with high-level semantics.
Despite its strong performance in image reconstruction, its main limitation is that mapping tokens back to raw pixels requires a large generative model (\textit{i.e.}, total 2.3B), which limits inference speed.
Although we significantly reduce inference time by pre-compressing images with VQ-VAE and simplifying inference steps via a one-step refiner, further efficiency gains for integrating with VLMs will require exploring lightweight models or techniques.

Moreover, due to limitations in computational resources and time, we only evaluate SemTok on image reconstruction and class-conditioned image generation tasks. 
Further validation is needed to assess its scalability and robustness across more diverse tasks, such as visual understanding, text-conditioned image generation, and video modeling.
For fair comparison, SemTok is trained only on ImageNet, inheriting its dataset bias, which limits qualitative comparison with approaches trained on larger-scale or commercial datasets.

\section{More Visualization Results}
\begin{figure*}[t]
\centering
\includegraphics[width=\linewidth]{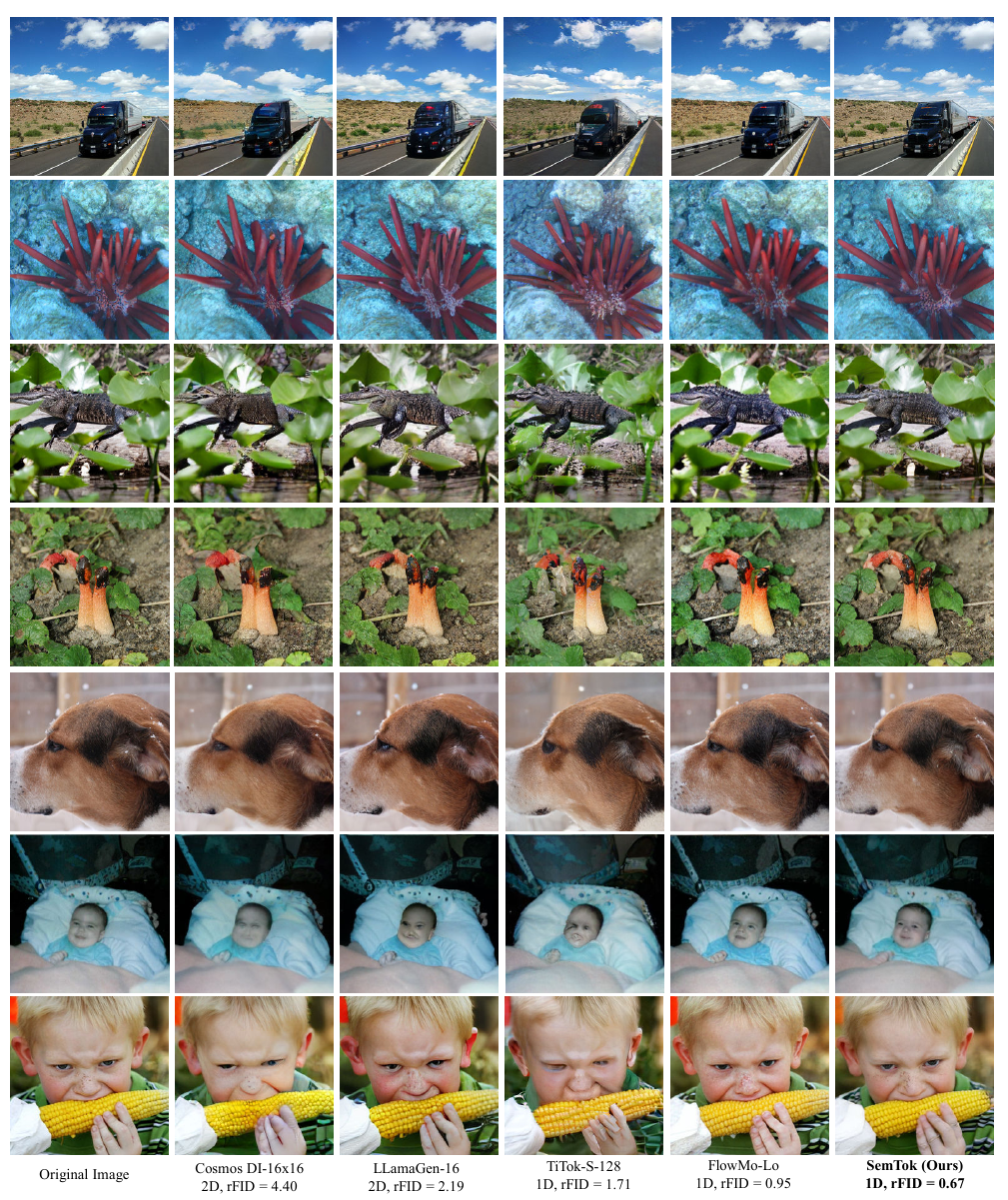}
\caption{Comparison of reconstructions from different tokenizers.}   
\label{fig:add_recon}
\end{figure*}

\begin{figure*}[t]
\centering
\includegraphics[width=\linewidth]{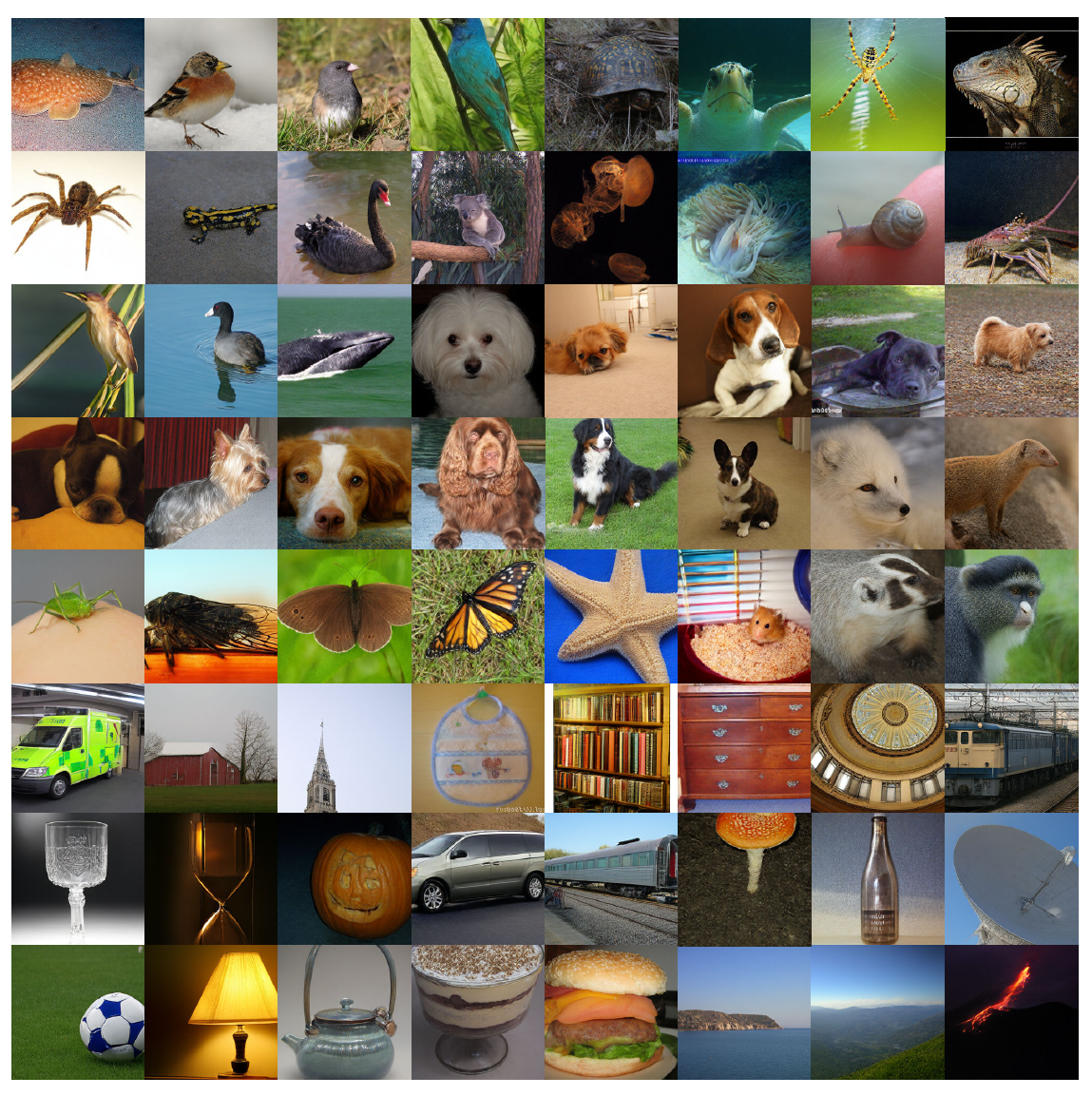}
\caption{Visualization of generated images from our SemTok-based AR model across random
ImageNet classes.}   
\label{fig:add_gen}
\end{figure*}

In Fig. \ref{fig:add_recon}, we provide extended comparisons with other tokenizers on the image reconstruction task, showcasing the robust capability of SemTok in reconstructing high-fidelity images.
In Fig. \ref{fig:add_gen}, we provide additional visualization results of our SemTok-based AR model on the image generation task, showcasing the robust capability of SemTok in generating high-realistic images.

\clearpage

%%%%%%%%%%%%%%%%%%%%%%%%%%%%%%%%%%%%%%%%%%%%%%%%%%%%%%%%%%%%%%%%%%%%%%%%%%%%%%%
%%%%%%%%%%%%%%%%%%%%%%%%%%%%%%%%%%%%%%%%%%%%%%%%%%%%%%%%%%%%%%%%%%%%%%%%%%%%%%%

\end{document}